\documentclass[aos]{imsart}

\usepackage{times,amsmath,amsfonts,amssymb,eucal,wasysym,mathrsfs,amsthm,mathtools,thmtools}
\usepackage{bbold,bm}
\usepackage{mathrsfs}
\RequirePackage[numbers]{natbib}
\RequirePackage[colorlinks,citecolor=blue,urlcolor=blue]{hyperref}
\RequirePackage{graphicx}
\usepackage{algorithm}
\usepackage{algorithmic}
\usepackage[capitalize,nameinlink]{cleveref}

\arxiv{2010.00000}
\startlocaldefs
\theoremstyle{plain}

\newtheorem{theorem}{Theorem}[section]
\newtheorem{lemma}[theorem]{Lemma}
\theoremstyle{definition}
\newtheorem{definition}[theorem]{Definition}

\theoremstyle{remark}

\usepackage[T1]{fontenc}    
\DeclareFontFamily{OT1}{pzc}{}
\DeclareFontShape{OT1}{pzc}{m}{it}{<-> s * [1.200] pzcmi7t}{}
\DeclareMathAlphabet{\mathpzc}{OT1}{pzc}{m}{it}
\usepackage[utf8]{inputenc} 

\def\a{\alpha}
\def\As#1{A^{(S)}}

\def\Acomp#1{A^{* #1}}

\def\b{{\beta}}
\def\Bs#1{B^{(S)}}

\def\d{\delta}
\def\diam{\text{diam}}
\def\D{\Delta}
\def\Ds{\tilde{\Delta}}
\def\Drg{D_{\text{rg}}}
\def\Dvp{D_{\text{vp}}}

\def\e{\varepsilon}

\def\f{{\bf f}}
\def\fs{\f^{(\ss)}}
\def\g{\gamma}
\def\grad{{\nabla}}
\def\l{\lambda}

\def\rp{\tilde{r}}
\def\s{\sigma}

\def\ts{\tilde{\theta}}
\def\that{\hat{\theta}}

\def\x{{\bf x}}
\def\xss{\x^{(\ss)}}
\def\xs#1{x^{(#1)}}

\def\xbs#1{\x^{(#1)}}

\def\xnots#1{\x^{\setminus(#1)}}
\def\X{{\bf X}}
\def\Xs#1{X^{(#1)}}
\def\Xbs#1{\X^{(#1)}}
\def\Xnots#1{\X^{\setminus(#1)}}
\def\xnots#1{\x^{\setminus(#1)}}

\def\xis#1{{\xi_{#1}^{*}}}
\def\xiS#1#2{{\xi_{#1}^{* #2}}}

\def\xp{\tilde{\x}}

\def\Xp{\tilde{\X}}

\def\XpS{\Xp^{(S)}}

\def\XnotpS{\Xp^{\setminus(S)}}

\def\Var{\text{Var}}

\def\yp{\tilde{y}}

\def\z{\zeta}

\def\Zn{Z^{(n)}}

\def\Zs{\tilde{Z}}
\def\znots{\z^{\setminus(S)}}

\def\E{\mathbb{E}}

\def\PP{\mathbb{P}}

\def\RR{\mathbb{R}}

\def\aa{\mathpzc{A}}
\def\bb{\mathpzc{B}}

\def\nn{\mathpzc{N}}

\def\ss{\mathpzc{S}}

\def\xx{\mathpzc{X}}
\def\yy{\mathpzc{Y}}

\def\core{{\ttfamily CORElearn}}

\def\knockoff{{\ttfamily knockoff}}
\def\gbm{{\ttfamily gbm}}
\def\glmnet{{\ttfamily glmnet}}
\def\rfsrc{{\ttfamily randomForestSRC}}
\def\varPro{{\ttfamily varPro}}

\def\ranger{{\ttfamily ranger}}

\DeclareFontFamily{U}{matha}{\hyphenchar\font45}
\DeclareFontShape{U}{matha}{m}{n}{
      <5> <6> <7> <8> <9> <10> gen * matha
      <10.95> matha10 <12> <14.4> <17.28> <20.74> <24.88> matha12
      }{}
\DeclareSymbolFont{matha}{U}{matha}{m}{n}

\DeclareMathSymbol{\Lt}{3}{matha}{"CE}
\DeclareMathSymbol{\Gt}{3}{matha}{"CF}

\def\({\left(}
\def\){\right)}
\def\[{\left[}
\def\]{\right]}

\def\bct{\begin{center}}
\def\ect{\end{center}}
\def\Array{\begin{eqnarray*}}
\def\EndArray{\end{eqnarray*}}
\def\Enumerate{\begin{enumerate}}
\def\EndEnumerate{\end{enumerate}}
\def\Eq{\begin{equation}}
\def\EndEq{\end{equation}}
\def\EqArray{\begin{eqnarray}}
\def\EndEqArray{\end{eqnarray}}
\def\Itemize{\begin{itemize}}
\def\EndItemize{\end{itemize}}
\def\mref#1{(\ref{#1})}
\def\qt#1{\qquad\text{#1}}

\def\Tabular{\begin{tabular}}
\def\EndTabular{\end{tabular}}
\def\FlushLeft{\begin{flushleft}}
\def\EndFlushLeft{\end{flushleft}}

\def\argmax{\mathop{\rm argmax}}

\def\equald{\,\smash{\mathop{=}\limits^{d}}\,}

\def\prob{\stackrel{\text{p}}\rightarrow}
\def\wp1{\stackrel{\text{with probability one}}\rightarrow}

\def\bb{\mathpzc{B}}

\def\ss{\mathpzc{S}}
\def\ss{\mathpzc{S}}
\def\xx{\mathpzc{X}}
\def\yy{\mathpzc{Y}}

\def\E{\mathbb{E}}
\def\PP{\mathbb{P}}
\def\RR{\mathbb{R}}

\def\a{\alpha}\def\b{\beta}\def\d{\delta}\def\D{\Delta}
\def\e{\varepsilon}\def\g{\gamma}
\def\l{\lambda}
\def\s{\sigma}
\def\z{\zeta}

\graphicspath{{figures/}}

\endlocaldefs

\begin{document}
\begin{frontmatter}
\title{Model-independent variable selection via the rule-based variable priority}
\runtitle{Variable selection via the rule-based variable priority}

\begin{aug}
\author[A]{\fnms{Min}~\snm{Lu}\ead[label=e1]{m.lu6@umiami.edu}\orcid{0000-0002-1386-1315}}
\and
\author[A]{\fnms{Hemant}~\snm{Ishwaran}\ead[label=e2]{hishwaran@miami.edu}\orcid{0000-0003-2758-9647}}
\address[A]{Division of Biostatistics, University of Miami\printead[presep={,\ }]{e1,e2}}
\runauthor{Lu and Ishwaran}
\end{aug}

\begin{abstract}
While achieving high prediction accuracy is a fundamental goal in
machine learning, an equally important task is finding a small number
of features with high explanatory power. One popular selection
technique is permutation importance, which assesses a variable's
impact by measuring the change in prediction error after permuting the
variable. However, this can be problematic due to the need to create
artificial data, a problem shared by other methods as well. Another
problem is that variable selection methods can be limited by being
model-specific.  We introduce a new model-independent approach,
Variable Priority (VarPro), which works by utilizing rules without the
need to generate artificial data or evaluate prediction error. The
method is relatively easy to use, requiring only the calculation of
sample averages of simple statistics, and can be applied to many data
settings, including regression, classification, and survival. We
investigate the asymptotic properties of VarPro and show, among other
things, that VarPro has a consistent filtering property for noise
variables. Empirical studies using synthetic and real-world data show
the method achieves a balanced performance and compares favorably to
many state-of-the-art procedures currently used for variable
selection.
\end{abstract}



\end{frontmatter}

\section{Introduction}

Although many machine learning procedures are capable of modeling a
large number of variables and achieving high prediction accuracy,
finding a small number of features with near-equivalent explanatory
power is equally desirable. This allows researchers to identify which
variables play a prominent role in the problem setting, thus providing
insight into the underlying mechanism for what might otherwise be
considered a black box. In machine learning, variable selection is
often performed using variable importance, described by how much a
prediction model's accuracy depends on the information in each
feature~\citep{Breiman2001, Friedman2001, van2006statistical,
  Ishwaran2007, strobl2008conditional, Ishwaran2008,
  doksum2008nonparametric, gromping2009variable, genuer2010variable,
  louppe2013understanding, williamson2023general}. One of the most
popular methods is permutation importance, introduced by Leo Breiman
in his famous random forests paper~\citep{Breiman2001}.  To calculate
a variable's permutation importance, the given variable is randomly
permuted in the out-of-sample data (out-of-bag, or OOB, data) and the
permuted OOB data is dropped down a tree. OOB prediction error is then
calculated. The difference between this and the original OOB error
without permutation, averaged over all trees, is the importance of the
variable. The larger the permutation importance of a variable, the
more predictive the variable.

Because permutation importance is not part of the forest construction
and does not make use of the forest ensemble, it can be considered a
\textit{filtering (screening)} feature selection method, which refers
to selection procedures performed without using a final prediction
model. Filtering procedures are widely adopted in many settings,
particularly for dealing with ultra-high dimensional
data~\citep{fan2008sure, zhu2011model, draminski2018rmcfs,
  tong2022model, zhong2023model}. On the other hand, permutation
importance, and other types of prediction-based importance measures,
can also be used for \textit{embedded} selection~\citep[see, for
  example,][]{diaz2006gene}, which refers to feature selection
processes embedded in the learning phase. An important class of
embedded procedures is penalization methods like the
lasso~\citep{tibshirani1996regression}. Recently, there has been
significant attention given to developing variable importance measures
that can apply more generally across different types of learning
procedures within the framework of model
selection~\citep{wei2015variable, lei2018distribution,
  fisher2019all}. \cite{modelfree} discuss the difference between
model selection and variable selection. An example of the latter is
the \textit{wrapper} approach~\citep{kohavi1997wrappers} for feature
subset selection. The term wrapper generally refers to a generic
induction algorithm used as a black box to score the feature
subsets. Another method worth mentioning is
knockoffs~\citep{candes2018panning}, which is a screening method that
can be applied across different procedures and has the useful property
of preserving the false discovery rate.

In this paper, we take a broader approach in the spirit of these
latter methods. We call our proposed method VarPro, which refers to a
model-independent framework of variable priority. The term
\textit{model-independent} reflects borrowing from the best parts of
both model-dependent selection and model-free variable selection in
the literature. This is because we construct trees just like in
permutation importance and tree filtering methods, but our goal is
different: to obtain rules and regions of the feature space over which
to calculate our importance score, rather than using predicted
outcomes and prediction error to determine importance. In this paper,
we are interested in developing a consistent model-independent
rule-based variable selection procedure applicable across different
data settings.

Let $Y$ be the response variable and $\Xs{1},\ldots,\Xs{p}$ the set of
$p$ potential explanatory features.  We consider the setting where the
researcher is interested in the conditional distribution of the
response $Y$ given the features $\X=(\Xs{1},\ldots,\Xs{p})$.
Our goal is to identify the variables of importance for a given function of the
conditional distribution of $Y$ given $\X$.  We call the target of
interest $\psi(\X)=\E(g(Y)|\X)$, where $g$ is a prechosen function 
specific to the problem being studied.  Some examples
are given below:
\Enumerate
\item
{\it Regression.} Here
$\psi(\X)=\E(Y|\X)$ where $g(Y)=Y$ and
the
goal is determining variables affecting the conditional mean.
\item
{\it Classification.} For a categorical response with
categories $c_1,\dots,c_L$, interest could focus on the conditional
probability $\psi(\X)=\PP\{Y=c_l|\X\}$ for a specific category $c_l$, where
$g(Y)=I\{Y=c_l\}$. For example, in studying the presence, absence, or recurrence of
cancer, the researcher may focus on the recurrence of cancer to
study the hypothesis that the probability of recurrence depends on
certain features.
\item
{\it Time to event.} With survival analysis, the focus of interest can
be the survival function $\psi(\X)=\PP\{T^o>t|\X\}$, where $Y=T^o$ is
the survival time.  In this case, $g(T^o)=I\{T^o>t\}$.
\EndEnumerate

Since it is expected that $\psi$ will depend on a smaller subset of
the $p$ variables $\ss\subset\{1,\dots,p\}$, the task is to find the
minimal set $\ss$ for which this holds, which we call the ``signal
variables'', while simultaneously excluding the non-relevant
variables, which we call ``noise variables''.  To make this idea more
precise we provide the following definition.  Write
$\Xbs{S}=\{\Xs{j}\}_{j\in S}$ for the feature vector $\X$ restricted
to coordinates $j\in S$ and $\Xnots{S}=\{\Xs{j}\}_{j\not\in S}$ for
coordinates not in $S$.

\vskip10pt
\begin{definition}\label{signal.noise.definition}
{\it $\ss\subset\{1,\ldots,p\}$ is the set of signal variables if $\ss$ is
  the minimal set of coordinates that $\psi$ depends on.  Thus, $\ss$
  is the smallest subset of coordinates satisfying
  $\PP\{\psi(\X)=\psi(\Xbs{\ss})\}=1$ such that
  $\PP\{\psi(\Xbs{\ss})=\psi(\Xbs{S'})\}=0$ for every $S'$
  where $S\nsubseteq S'$.
  The complementary set
  $\nn=\{1,\ldots,p\}\setminus \ss$ is unrelated to $\psi$ and
  therefore contains the noise variables.  If
  $\ss=\emptyset$, then $\psi=\E(g(Y))$ is constant and
  $\nn=\{1,\ldots,p\}$. However, we rule this trivial case out
  and always assume $\ss\neq\emptyset$.}
\end{definition} 

The definition of $\ss$ specifies that it is the smallest set of
coordinates such that $\psi(\x) = \psi(\xbs{\ss})$ for almost all
$\x$. Furthermore, there is no other set $S' \ne \ss$ (where $\ss
\nsubseteq S'$) for which $\psi(\X) = \psi(\Xbs{S'})$ holds with
non-zero probability.  Another way to
view~\cref{signal.noise.definition} is that it implies a type of
conditional independence for noise variables.  For example, if $p=2$
and $\ss=\{1\}$, then since $\Xs{2}$ is a noise variable
\Eq
\E\(g(Y)|\Xs{1},\Xs{2}\)
=\E\(g(Y)|\Xs{1}\).
\label{signal.noise.example}
\EndEq
We note~\mref{signal.noise.example} is weaker than the usual
conditional independence assumption used for variable selection (see
equation~\mref{strong.conditional} below) since it depends upon the
choice of $g$ and $\psi$.  For example, in classification, if the
analysts chooses $g(Y)=I\{Y=c_l\}$ for a specific class label $c_l$ of
interest,~\mref{signal.noise.example} implies a conditional
independence of $\Xs{2}$ for class label $c_l$, but not necessarily
for other class labels.  Thus, variables affecting relapse of cancer
may be different from those affecting absence or death due to cancer.
The ability to select $(g,\psi)$ to suit the data question being
studied is an important feature of our approach and provides
researchers with a flexible tool to study how variables play a role in
their data settings.  As another example, consider survival analysis.
As discussed in Section 5, a useful way to summarize lifetime behavior
is by restricted mean survival time (RMST)~\citep{royston2011use}.
The RMST equals the integrated survival function up to a fixed time
horizon value $\tau>0$ and equals
$$
\psi_\tau(\X)=\int_0^\tau \PP\{T^o>t|\X\}\,dt
=\E(g(T^o)|\X),\qt{where } g(T^o)=T^o \wedge \tau.
$$
Because $\psi_\tau(\X)$ can vary with time, the list of signal
variables can vary with $\tau$.  For example, identifying variables
affecting year one breast cancer survival is crucial for properly
managing early treatment, but knowing which variables affect lifetime
after year one is equally important for tailoring treatment for
long-term survival.

Our approach works with the function $\psi$ obtained by integration
with respect to the conditional distribution of $Y$ given $\X$, but
another method often used is to work directly with the conditional
distribution. This assumes a conditional independence between $Y$ and
the noise variables given the signal variables:
\Eq
Y\perp \Xnots{\ss}|\Xbs{\ss}.
\label{strong.conditional}
\EndEq
Let $S \subset \{1,\ldots,p\}$ be the variables of interest, and
the goal is to determine if $S$ contains signal variables. The
strategy is to construct a test statistic $\that_n(S)$ using augmented
features $(\Xbs{S},\Xnots{S},\XpS,\XnotpS)$, where $\XpS$ and
$\XnotpS$ are new artificial features. The test is constructed in such
a way that, due to conditional independence, $\that_n(S)$ is
statistically non-significant if $S$ are noise variables.

We mention two popular methods using this idea. The first is
permutation importance for random forests mentioned earlier (referred
to hereafter as VIMP for variable importance). In VIMP, the feature
vector $\Xnots{S}$ is permuted to obtain $\XnotpS$; then the predicted
value for $(\Xbs{S},\Xnots{S})$ is compared to the predicted value for
$(\Xbs{S},\XnotpS)$, where this difference should be nearly zero
if~\mref{strong.conditional} holds. However, a well-known problem with
VIMP is that the permuted sample does not have the same distribution
as $\X$, which can lead to flawed variable
selection~\citep{strobl2007bias}.  The second technique is
knockoffs~\citep{candes2018panning}. In knockoffs, a simulation
according to the distribution of $\X$ is used to obtain
$(\XpS,\XnotpS)$, where the artificial data is simulated so as to
satisfy
\Eq
(\Xbs{S},\Xnots{S},\XpS,\XnotpS)\equald
(\Xbs{S},\XnotpS,\XpS,\Xnots{S}).
\label{knockoff}
\EndEq
This is used to compute a knockoff statistic for filtering
variables. By making use of~\mref{strong.conditional}, the knockoff
test statistic can achieve a desired false discovery level. This novel
idea avoids the problems of permutation importance; however, it may
rely on strong assumptions about the parent distribution, and
achieving~\mref{knockoff} could be difficult in certain situations.

\subsection{Contributions of this work and outline of the paper}

The VarPro method introduces a novel approach to using rules. A
similar method is the mean decrease impurity (MDI)
score~\citep{Breiman1984}. For a given variable $\Xs{s}$, the MDI
score is obtained by summing the weighted impurity decrease across all
tree nodes that involve a split using $\Xs{s}$~\citep[for forests, the
  MDI is averaged over all trees in the forest;
  see][]{louppe2013understanding}. This score provides a useful
summary of a feature's importance and is used for ranking
variables. When the Gini index is used as the impurity function in
classification, this measure is known as the Gini importance. VarPro
can be viewed as a generalization of MDI since MDI is approximately
recovered in the proposed framework when setting $g$ to either the
Gini index for classification or the variance in regression. However,
VarPro extends MDI because the VarPro importance score for a feature
is determined by a local estimator \textit{that drops all conditions
  on the feature at once}, whereas MDI considers the marginal effect
of the feature (i.e., the impurity drop at a specific node that splits
on $\Xs{s}$).

The structure of the paper is as follows. Section 2 describes the
VarPro procedure in detail. Given a rule, the VarPro importance score
for a set of variables $S$ is defined as the difference between two
local estimators of $\psi$, where one is estimated from the region of
the data defined by the rule and the other by removing any constraints
on the rule involving the features in $S$. The local estimators are
simple averages of statistics and are relatively easy to
calculate. Examples are provided to motivate the method and explain
the concepts of a region and its released region, which are essential
to understanding the VarPro importance score. We then contrast VarPro
with permutation importance to highlight its advantages. In this
discussion, we find that VarPro can be written as a permutation test,
allowing us to pinpoint the shortcomings of permutation importance and
demonstrate how VarPro overcomes these issues.

Section 3 provides theoretical justification for VarPro. Our theory
considers both the null case, when the variable is noise, and the
alternative case, when the variable has signal. We show that VarPro
consistently filters out noise variables and, for signal variables, we
derive the asymptotic limit of a bias term representing the limiting
VarPro importance score. These results hold under relatively mild
conditions, such as a smoothness property for $\psi$ and certain
conditions for the rules used by VarPro. These conditions are expected
to hold for any reasonably constructed tree procedure, making VarPro
generally agnostic to the rule-based procedure used for rule
generation.

Section 4 presents empirical results demonstrating VarPro's
effectiveness using synthetic and real-world data for regression and
classification in both low and high-dimensional feature
spaces. Building on the work in Section 2, we extend VarPro to
survival analysis in Section 5. A large cardiovascular study and
high-dimensional simulations are used to illustrate the effectiveness
of this extension. Section 6 concludes with a discussion summarizing
the strengths and weaknesses of the method. All proofs and
supplementary information are located in the Appendix.

\section{A new rule-based variable selection approach}

We begin by providing a broad overview of the idea, after which we
will delve into specific details. Let $S$ again represent the set of
variables of interest. Given a rule $\zeta$, VarPro calculates a
sample averaged estimator $\that_n(\zeta)$ for the target function
$\psi$ of interest by using the data in $\zeta$'s
\textit{region}. Then a \textit{released rule} $\zeta^S$ is
constructed by removing any constraints on the indices in $S$, and its
sample averaged estimator $\that_n(\zeta^S)$ is calculated over the
\textit{released region}. The estimator $\that_n(\zeta^S)$ is then
contrasted with $\that_n(\zeta)$ to measure the importance of $S$.
Many existing methods for variable importance rely on either
resampling or refitting models, which can introduce finite sample
bias, or they make use of artificial data as described earlier. A
unique feature of VarPro is that it bypasses the need for this, and
instead constructs estimators $\that_n(\zeta)$ and $\that_n(\zeta^S)$
directly from the data. These estimators serve as local estimates of
$\psi$, and because $\that_n(\zeta^S)$ is calculated using the data
from releasing coordinates on $S$, it implies under certain conditions
due to the laws of averages that $|\that_n(\zeta) - \that_n(\zeta^S)|
\prob 0$ for noise variables $S$. However, for signal variables, the
limit is different. Therefore, this makes it possible to consistently
filter noise variables using the VarPro importance score
$|\that_n(\zeta) - \that_n(\zeta^S)|$.

\subsection{Regions and release regions}

A key aspect of VarPro is the idea of the region of a rule $\z$.
A region is a subset of
the feature space obtained by a function $R$ that maps $\z\mapsto
\xx$.  In many cases, $\z$ can be associated with a series of
univariate rules.  The region mapped by $R$ in this case is
denoted by $R(\z)=\{\x\in\xx: \xs{1}\in R(\z^{\Xs{1}}), \ldots, \xs{p}\in
R(\z^{\Xs{p}})\}$, where $\z^{\Xs{j}}$ denotes the univariate rule for
feature $j$.  For example, if all the features are continuous, we
can imagine a rule with region $R(\z)=\{\x\in \RR^p: a_1\le
\xs{1}\le b_1,\ldots, a_p\le \xs{p}\le b_p\}$.
Boundaries like this naturally
arise in machine learning methods constructed from decision rules.

In order to construct the VarPro importance score we need to
introduce the idea of a released rule and a released region.

\begin{definition} \label{releasedRule}
{\it Let $R(\z)\subset\xx$ be a region of the feature space.  To check the
importance of the variables $\Xbs{S}$ to the rule $\z$, we introduce
the concept of a released region $R(\z^S)$ for the released rule $\z^S$
obtained by removing the dependence of $\z$ on the coordinates
$\Xbs{S}$:
$$
R(\z^S) = \{\x\in\xx: \xnots{S}\in R(\z)\}.
$$
In other words,  the released region $R(\z^S)$ is the
set of all $\x$ whose $S$-coordinate values are unconstrained
but with non-$S$-coordinate values that lie in $R(\z)$.  In
particular, $R(\z)\subseteq R(\z^S)$.}
\end{definition}

\begin{figure}[phtb]
  \vspace{1em}
  \makebox[\textwidth][c]{%
    \makebox[0.4\textwidth]{\centering (A) Original Rule}%
    \hspace*{5em}
    \makebox[0.4\textwidth]{\centering (B) Permutation Variable Importance (VIMP)}%
  }
  \hspace*{-.25in}\scalebox{1.25}{\includegraphics{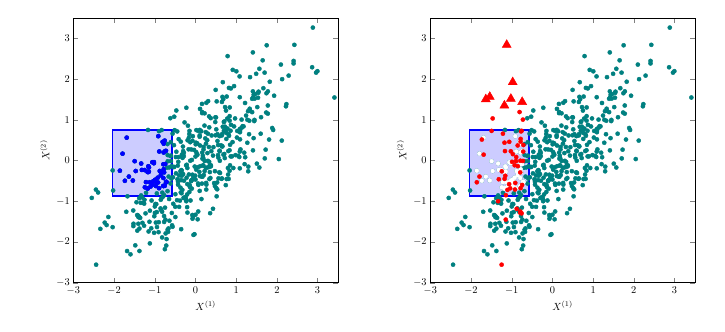}}
  \\[10pt]
  \vspace{1em}
  \makebox[\textwidth][c]{%
    \makebox[0.4\textwidth]{\centering (C) VarPro Release Region}%
    \hspace*{2em}
    \makebox[0.4\textwidth]{\centering (D) VarPro Importance Score}%
  }
  \hspace*{-.25in}\scalebox{1.25}{\includegraphics{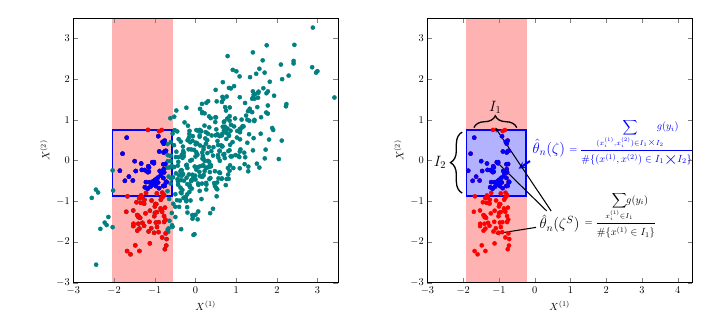}}
\caption{\small Two-dimensional illustration of how VarPro differs from
  artificial data methods.  (A) The two-dimensional region for $\z$ is
  a rectangle.  The data of interest are marked in blue.
  (B) Permutation variable importance (VIMP) for $\Xs{2}$.  The data
  was permuted along $\Xs{2}$ and data
  marked in red with triangles identify values that
  do not match the joint distribution of $\X$.  The model-based
  predicted values $\yp:=\yp(\xs{1},\tilde{x}^{(2)})$ for these
  artificial points are extrapolated from a region of the feature
  space that could be from potentially different responses.
  (C) VarPro release region for
  $\Xs{2}$. The original rule is modified to the $S$-released rule
  $\z^S$ (where $S=\{2\}$) shown using a pink background color.
  (D) VarPro importance score is defined using the
  estimator calculated using observed data values in blue compared to
  the estimator where the new released values in red are additionally
  used.  No  artificial data needs to be created. }
\label{illus1}
\end{figure}

{\bf Example 1.}  In this example, shown in~\cref{illus1}, the feature
space is $\xx \subseteq \RR^2$ and the rule $\zeta$ corresponds to the
branch of a tree using splits $\xs{1} \leq -0.7$, $\xs{2} \geq -0.8$,
$\xs{2} \leq 0.7$, and $\xs{1} \geq -1.95$. We can think of $\zeta$ as
a product of indicator functions:
$$
\zeta =
I\{\xs{1} \leq -0.7\} \cdot
I\{\xs{2} \geq -0.8\} \cdot
I\{\xs{2} \leq 0.7\} \cdot
I\{\xs{1} \geq -1.95\}.
$$
The region $R(\zeta)$ is the blue rectangle in~\cref{illus1} (A):
$$
R(\zeta) = \{(\xs{1}, \xs{2}): -1.95 \leq \xs{1} \leq -0.7, -0.8 \leq \xs{2} \leq 0.7\}.
$$
Consider testing whether $\Xs{2}$ is a noise variable. Then, by~\cref{releasedRule}, we set $S = \{2\}$ and obtain the released region by removing the dependence on coordinate $\Xs{2}$:
$$
R(\zeta^S) = \{(\xs{1}, \xs{2}): -1.95 \leq \xs{1} \leq -0.7\}.
$$
The released region is therefore the rectangle with the sides for the
released coordinates removed (see the pink region in~\cref{illus1}
(C)). Interestingly, since we are working with a classical tree, the
released rule is itself a tree branch equal to the original rule
altered such that whenever a binary decision is made on a variable in
$S$, the decision is always 1. In this example, $\zeta^S = I\{\xs{1}
\leq -0.7\} \cdot I\{\xs{1} \geq -1.95\}.$
\qed

\begin{figure}[phtb] 
 \centering
 \resizebox{6in}{!}{\includegraphics[page=1]{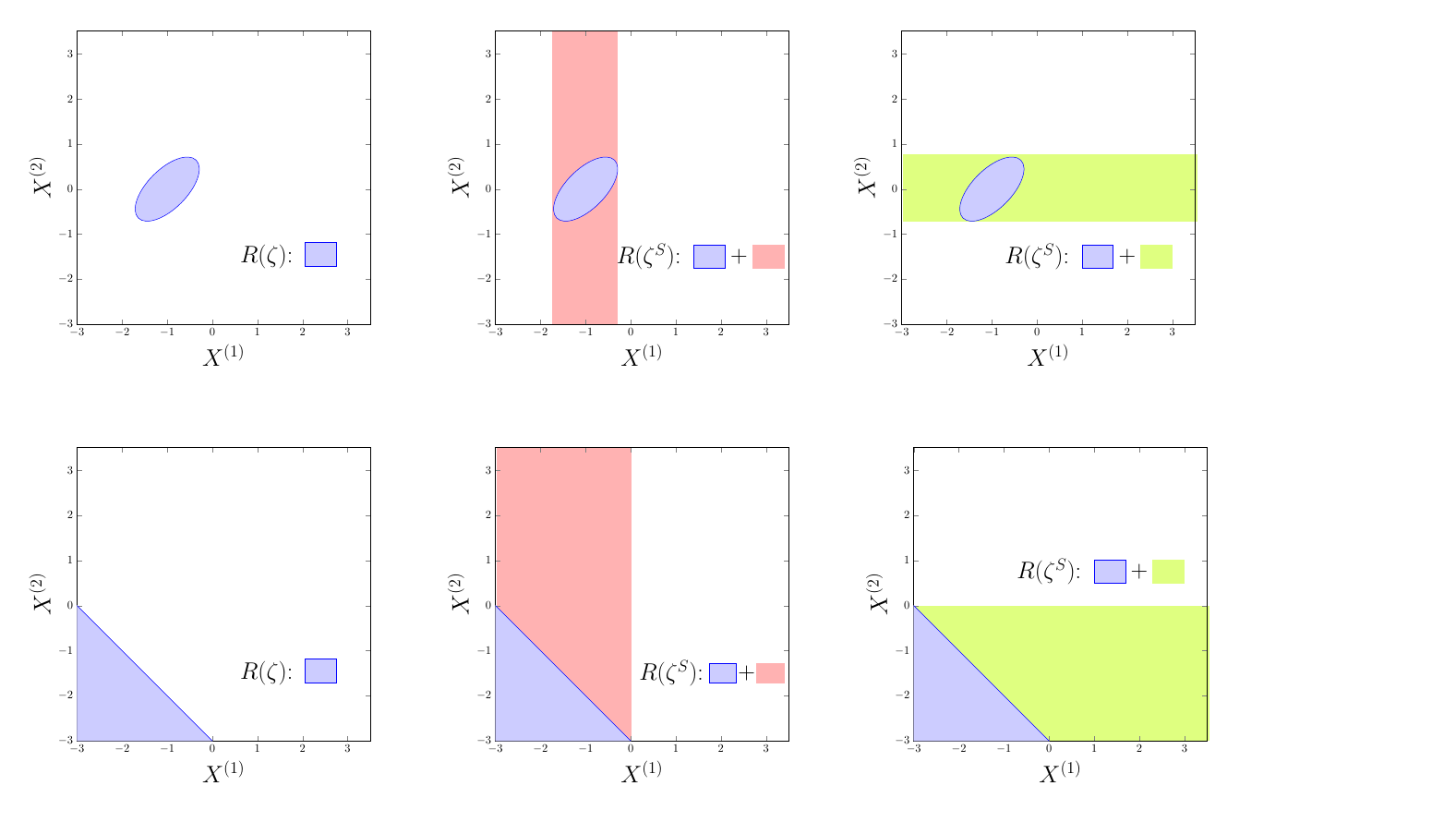}}
\vskip-10pt
\caption{\small Regions $R(\z)$ (in blue) for rules $\z$ produced by
  a machine learning procedure.
  Top left is for an elliptical rule; bottom left 
  is for a hyperplane rule.  Middle and right column figures are release
  region $R(\z^S)$ when releasing coordinates $\Xs{2}$ with $S=\{2\}$
  (red plus blue region) and $\Xs{1}$ with $S=\{1\}$ (green plus
  blue region), respectively.}
\label{releaseRegions}
\end{figure}

\vskip15pt
{\bf Example 2.}
\cref{releaseRegions} illustrates how~\cref{releasedRule} can apply to
rules other than those from classical trees. In both illustrations, the
region of interest is a function of the variables that is not
decomposable into products of conditions on individual variables as in
a typical tree. The top left panel shows an elliptical region, which
is a rule condition given by a quadratic inequality relating to both
variables. According to~\cref{releasedRule}, to release the region
along $\Xs{2}$, we take the set of all $\x$ with coordinates $\Xs{1}$
inside the ellipse, while $\Xs{2}$ is left unconstrained. Therefore,
the released region $R(\z^S)$ is a box that touches the left and right
sides of the ellipse as shown in the top middle panel. The panel to
its right shows the release region when $\Xs{1}$ is released. This is
a box that touches the top and bottom of the ellipse. The second
illustration, given in the bottom left of~\cref{releaseRegions}, is a
hyperplane region. This is a rule condition given by a simple linear
inequality condition relating to both variables (for instance, this
would occur for rules extracted from random projection trees). The
middle and right bottom panels display the release region when
releasing coordinates $\Xs{2}$ and $\Xs{1}$.
Observe that in both of our examples, the released rule $\z^S$
obtained by releasing $S$ may not correspond to a rule generated by
the original machine learning procedure. This is because if the region
is a function that depends on the features using some type of
threshold function, then the resulting rule may be different from the
rules generated by the original procedure. Nonetheless, our framework
is designed to accommodate such a scenario.
\qed

\vskip10pt

\subsection{VarPro importance score}

The VarPro importance score is defined formally as follows.
Let $(\X_1,Y_1),\ldots,(\X_n,Y_n)\in\xx\bigtimes\yy$ be the
i.i.d.~learning data sampled from $\PP$.  Observations $Y_i$ for
$\X_i\in R(\z)$ are used to estimate the conditional mean of $g(Y)$ in
$R(\z)$ via
\Eq
\that_n(\z)=\frac{1}{\#\{\X_i\in R(\z)\}}\sum_{\X_i\in
  R(\z)} g(Y_i).
\label{varpro.estimator}
\EndEq
The released region $R(\z^S)$ is used to obtain the released estimator
\Eq
\that_n(\z^S)=\frac{1}{\#\{\X_i\in R(\z^S)\}}\sum_{\X_i\in R(\z^S)} g(Y_i).
\label{varpro.released.estimator}
\EndEq
The VarPro importance of $S$ is defined as the
absolute difference between~\mref{varpro.estimator}
and~\mref{varpro.released.estimator},
$|\that_n(\z^S)-\that_n(\z)|$.

We return to our previous example of~\cref{illus1} to give some
intuition for this estimator by contrasting it to
permutation importance (VIMP).  Recall that $R(\z)$ is the blue
rectangle in (A) and the released region $R(\z^S)$ is the pink area in
(C).  Panel (B) displays the data after permutation on coordinate
$\Xs{2}$.  Permuted data $(\xs{1},\tilde{x}^{(2)})\in R(\z^S)$ are
marked in red.  Notice that in some cases (indicated by
triangles) these data deviate strongly from the true distribution of
$(\Xs{1},\Xs{2})$.  This creates problems for VIMP. This is because
VIMP calculates the importance of $\Xs{2}$ by comparing the test
statistic calculated using the observed data in $R(\z)$ (blue points
in (A)) to that calculated using the permuted data in $R(\z^S)$ (red
points in (B)).  If this were a regression setting, 
this corresponds to averaging
observations $y_i$ for $i\in R(\z)$ and comparing them to averaged
estimated values $\yp_i:=\yp_i(\xs{1}_i,\tilde{x}^{(2)}_i)$.
Unfortunately, since $\yp_i$ has to be model estimated (since it
must estimate $\psi$ for features not in the training data), this
produces atypical $\yp_i$ if $\tilde{x}^{(2)}_i$ is from a
different region of the data space than the original data (like the red
triangular points).  This can result in large VIMP even when $\Xs{2}$
is a noise variable.

This problem occurs because the learner is being applied to
artificially created data that deviates from the feature
distribution.  To avoid this, VarPro takes a direct approach.
Using the original training
data, estimates are obtained for the conditional mean of $g(Y)$ for
$\X\in R(\z)$
using observed values
$g(y_i)$ via~\mref{varpro.estimator}
and~\mref{varpro.released.estimator}, where in a
regression setting 
$g(y)=y$.  Shown in~\cref{illus1} (D), $\that_n(\z)$ is calculated from
observations in the blue box (blue points), which is compared to
$\that_n(\z^S)$ calculated using the data in the pink rectangle (blue
and red points), yielding the importance value
$$
\Bigg|\that_n(\z^S)-
\that_n(\z)\Bigg|
=\left|\frac{\displaystyle{\sum_{\xs{1}_i\in I_1} y_i}}{\#\{\xs{1}\in I_1\}}
-\frac{\displaystyle{\vphantom{\sum_{\xs{1}_i\in I_1} y_i}
    \sum_{(\xs{1}_i,\xs{2}_i)\in I_1\bigtimes I_2} y_i}}
{\vphantom{\#\{\xs{1}\in I_1\}}\#\{(\xs{1},\xs{2})\in I_1\bigtimes I_2\}}\right|,
$$
where $I_1=[-1.95,\,-0.7]$ and $I_2=[-.8,\,0.7]$.  If $\Xs{1}$ is a signal
feature, but $\Xs{2}$ is a noise variable, then the two averages should
converge to nearly the same value because $\psi(\X)=\E(Y|\X)=\psi(\Xs{1})$
depends only on $\Xs{1}$.  On the other hand, if $\Xs{2}$ is a signal
feature, then the difference will not necessarily be zero.

\vskip10pt{\scshape VarPro Key Idea.}\emph{ Releasing a rule $\z$ along 
the noise coordinates does not change $\psi$,
therefore the difference between the average of $g$ in the
rule's region and its released region will be
asymptotically the same, and subsequently we can expect a zero
importance.  For signal variables, the opposite happens, and since
$\psi$ changes along the released direction, we
can expect a difference in average $g$ values in the released and
non-released regions, and therefore we can expect a non-zero
importance value asymptotically.}
\vskip2pt

\subsection{Connection of VarPro to permutation tests}

We use the
following construction to gain insight into the dangers of using
artificial data as well as
to expand on our previous points about VIMP.
Surprisingly, it turns out that we can write the VarPro
estimator as a specialized type of permutation test.
This will be
contrasted against  model-based importance to explain
what goes wrong with artificial data and to explain how VarPro
avoids this.  This also motivates the idea of using an external
estimator to be used in Section 5
for survival analysis.

Without loss of generality, assume that $\x$ is
reordered so that $\x=(\xbs{S},\xnots{S})$.
Consider all possible permutations of the training data
$\{(y_i,\xbs{S}_j,\xnots{S}_i)\}_{1\le i,j\le n}$.  Then it will be
shown (see below) that
\Eq
\frac{\sum_{i=1}^n g(y_i)I\{\x_i\in R(\z^S)\}}{\sum_{i=1}^n I\{\x_i\in R(\z^S)\}}
=
\frac{\sum_{i=1}^n\sum_{j=1}^n g(y_i)I\{(\xbs{S}_j,\xnots{S}_i)\in R(\z)\}}
     {\sum_{i=1}^n\sum_{j=1}^n I\{(\xbs{S}_j,\xnots{S}_i)\in R(\z)\}}.
\label{varpro.permute}
\EndEq
The left-hand side is the VarPro released estimator $\that_n(\z^S)$ described
in~\mref{varpro.released.estimator}, whereas the right-hand side is an
estimator using the permuted data.
Obviously in practice the permuted estimator on the right side
of~\mref{varpro.permute} is impractical since it involves $O(n^2)$
calculations and would be an inefficient way to calculate
$\that_n(\z^S)$.  However, we use it to describe why
permutation VIMP generally does not satisfy an identity
like~\mref{varpro.permute}, which will highlight the problem.

First, however, let us explain why~\mref{varpro.permute} holds.  Let
$\znots$ be the rule that relaxes the constraints of $\z$ on the
indices not in $S$.  We have the following identity for rules expressible
as products $\z = \prod_{s=1}^p
I\{\xs{s}\in I_s\}$ where $I_s\subseteq\RR$ are real-valued intervals
(this is like the example described in~\cref{illus1}):
\Eq
I\{\x\in R(\z)\}
=I\{\xnots{S}\in R(\z)\}\cdot I\{\xbs{S}\in R(\z)\}
=I\{\x\in R(\z^S)\}\cdot I\{\x\in R(\znots)\}.
\label{varpro.permute.assumption}
\EndEq
The right side follows by \cref{releasedRule} 
since $\x\in R(\z^S)$ implies  $\xnots{S}\in R(\z)$ and
$\x\in R(\znots)$ implies $\xbs{S}\in R(\z)$ and vice-versa.

Therefore, for rules satisfying~\mref{varpro.permute.assumption} 
\EqArray
I\{(\xbs{S}_j,\xnots{S}_i)\in R(\z)\}
&=&
I\{(\xbs{S}_j,\xnots{S}_i)\in R(\z^S)\}\cdot
I\{(\xbs{S}_j,\xnots{S}_i)\in R(\znots)\}\nonumber\\
&=&
I\{\x_i\in R(\z^S)\}\cdot
I\{\x_j\in R(\znots)\}
\label{varpro.permute.identity}
\EndEqArray
where the last line is because $(\xbs{S}_j,\xnots{S}_i)\in R(\z^S)$
depends only on $\xnots{S}_i$ and $(\xbs{S}_j,\xnots{S}_i)\in
R(\znots)$ depends only on $\xbs{S}_j$.
Hence we have
\Array
&&\hskip-35pt
\frac{\sum_{i=1}^n\sum_{j=1}^n g(y_i)I\{(\xbs{S}_j,\xnots{S}_i)\in R(\z)\}}
     {\sum_{i=1}^n\sum_{j=1}^n I\{(\xbs{S}_j,\xnots{S}_i)\in R(\z)\}}\\
&&\qquad=
     \frac{\sum_{i=1}^n\sum_{j=1}^n g(y_i)
       I\{\x_i\in R(\z^S)\}
       I\{\x_j\in R(\znots)}
     {\sum_{i=1}^n\sum_{j=1}^n 
       I\{\x_i\in R(\z^S)\}
       I\{\x_j\in R(\znots)\}}\\
&&\qquad=
     \frac{\sum_{i=1}^n g(y_i)
       I\{\x_i\in R(\z^S)\} \sum_{j=1}^n I\{\x_j\in R(\znots)\}}
     {\sum_{i=1}^n I\{\x_i\in R(\z^S)\}
       \sum_{j=1}^n I\{\x_j\in R(\znots)\}}\\
&&\qquad=
    \frac{\sum_{i=1}^n g(y_i)I\{\x_i\in R(\z^S)\}}{\sum_{i=1}^n
      I\{\x_i\in R(\z^S)\}}:=\that_n(\z^S).
\EndArray
The first line is from~\mref{varpro.permute.identity}.
The last line is due to the cancellation of the common term in
the numerator and denominator which is directly related to working
with $g(y_i)$.  

Now let us compare this to 
VIMP.
Recall that unlike VarPro, VIMP does not use the observed
response to estimate $\psi$ but uses a model-based
estimator instead.  Let $\psi_n$ be this estimator.  For
example, this could be the ensemble estimator from a random forest
analysis or a tree boosted estimator using gradient boosting.
Then using the permuted data as above, the VIMP estimator
is
\Array
\ts_{\text{VIMP}}(\z^S)
&=&\frac{\sum_{i=1}^n\sum_{j=1}^n \psi_n(\xbs{S}_j,\xnots{S}_i)
   I\{(\xbs{S}_j,\xnots{S}_i)\in R(\z)\}}
{\sum_{i=1}^n\sum_{j=1}^n I\{(\xbs{S}_j,\xnots{S}_i)\in R(\z)\}}\\
&=&
\frac{\sum_{i=1}^n\sum_{j=1}^n \psi_n(\xbs{S}_j,\xnots{S}_i)
  I\{\x_i\in R(\z^S)\} I\{\x_j\in R(\znots)\}}
{\sum_{i=1}^n I\{\x_i\in R(\z^S)\} \sum_{j=1}^n I\{\x_j\in R(\znots)\}}
\EndArray
where the last identity is due to~\mref{varpro.permute.identity}.

Notice, however, that the cancellation that occured previously in the
numerator and denominator is no longer guaranteed to hold, and it is
not true that $\ts_{\text{VIMP}}(\z^S)$ is the same as the
non-permuted estimator
\Eq
\ts_n(\z^S)
=\frac{\sum_{i=1}^n \psi_n(\x_i) I\{\x_i\in R(\z^s)\}}
{\sum_{i=1}^n I\{\x_i\in R(\z^s)\}},
\label{vimp.nopermute}
\EndEq
which is the analog to $\that_n(\z^S)$ since it replaces
the observed value $g(y_i)$ with the model estimated value
$\psi_n(\x_i)$.  
Just like $\that_n(\z^S)$, the estimator $\ts_n(\z^S)$ leads to 
consistent variable selection (to be shown
in~\cref{varpro.external.noise.theorem}) so the equality would show that
the model-based permutation importance has good properties.   But
this cancellation occurs and the two estimators become the same only
if
\Eq
\psi_n(\x)=\psi_n(\xnots{S})
\label{vimp.permute.assumption}
\EndEq
because then 
\Array
\ts_{\text{VIMP}}(\z^S)
&=&
\frac{\sum_{i=1}^n \psi_n(\xnots{S}_i)
  I\{\x_i\in R(\z^S)\} \sum_{j=1}^n I\{\x_j\in R(\znots)\}}
{\sum_{i=1}^n I\{\x_i\in R(\z^S)\} \sum_{j=1}^n I\{\x_j\in R(\znots)\}}\\
&=&
\frac{\sum_{i=1}^n \psi_n(\xnots{S}_i) I\{\x_i\in R(\z^S)\}}
{\sum_{i=1}^n I\{\x_i\in R(\z^S)\}}:= \ts_n(\z^S).
\EndArray
However~\mref{vimp.permute.assumption} is a very strong assumption.
If $S$ consists entirely of noise variables then
$\psi(\x)=\psi(\xnots{S})$, so in this
case~\mref{vimp.permute.assumption} is asserting that the model-based
estimator has correctly eliminated all noise variables.  This is a
very strong requirement and is asking a lot from the underlying
procedure since after all this would mean that the procedure has
achieved perfect variable selection on its own.  Therefore, it is not
reasonable to expect~\mref{vimp.permute.assumption} to hold in general, which
means $\ts_{\text{VIMP}}(\z^S)$ will not equal $\ts_n(\z^S)$ in
general.

We emphasize that while this highlights the issue of using $\psi_n$
with $\ts_{\text{VIMP}}(\z^S)$, this does not necessarily detract from
the potential usefulness of an external estimator. A point we explore
is how to use the external estimator in a more coherent way. We argue
that the non-permuted estimator $\ts_n(\z^S)$ defined
in~\mref{vimp.nopermute} presents one such opportunity. After all, we
know that machine learning methods like gradient boosting and random
forests produce highly accurate estimators in many settings. VIMP
misuses $\psi_n$ by applying it to artificial data, which can produce
biased estimation. However, the non-permuted estimator $\ts_n(\z^S)$,
which was introduced as an analog to the VarPro estimator, only
applies $\psi_n$ to the training data. This is why it does not equal
permutation VIMP and why it represents a legitimate way to proceed.

In fact, in Section 5 we develop an estimator like this for
survival analysis to extend VarPro to handle censored data.
As will be discussed,
in censored settings an external estimator makes sense.  However at
the same time,
if the response is
observed, then we prefer to use the VarPro importance score
$$
\Big|\that_n(\z^S)-\that_n(\z)\Big|
=\Bigg|
\frac{\sum_{i=1}^n g(y_i)I\{\x_i\in R(\z^S)\}}{\sum_{i=1}^n
  I\{\x_i\in R(\z^S)\}}
-
\frac{\sum_{i=1}^n g(y_i)I\{\x_i\in R(\z)\}}{\sum_{i=1}^n
  I\{\x_i\in R(\z)\}}\Bigg|
$$
which is model-independent and makes use of the actual observed responses, rather than 
an external importance score
$$
\Big|\ts_n(\z^S)-\ts_n(\z)\Big|
=\Bigg|
\frac{\sum_{i=1}^n \psi_n(\x_i)I\{\x_i\in R(\z^S)\}}{\sum_{i=1}^n
  I\{\x_i\in R(\z^S)\}}
-
\frac{\sum_{i=1}^n \psi_n(\x_i)I\{\x_i\in R(\z)\}}{\sum_{i=1}^n
  I\{\x_i\in R(\z)\}}\Bigg|
$$
which is model-specific.  The former leads to a simpler procedure
based on averages of independent observations which is consistent
under relatively mild conditions.  Also it is easier to apply to
different $g$ functions without having to fit a new learning method
when $g$ is changed by the researcher.

\section{Theory}

How does one construct a rule $\z$ to identify variables informative
for $\psi$?  In practice, there are many procedures to
choose from that produce rules for VarPro, including
simple decision rules~\citep{tan2005simple}, rule
learning~\citep{furnkranz1997pruning}, trees~\citep{Breiman1984},
Bayesian trees~\citep{GMT}, Bayesian additive regression
trees~\citep{chipman2010bart}, Bayesian forests~\citep{Baysianforest}
and random forests~\citep{Breiman2001}.  In the examples presented in
this paper, the rules are chosen by randomly selecting branches from a
CART tree.

We assume hereafter that we have constructed a rule, or more
generally, a collection of rules for a specific problem.  Previously
we had defined the VarPro importance score
on the basis of a single rule, but
we will actually use many rules and average their scores
to obtain a more stable estimator (see the discussion
following~\cref{varpro.signal.theorem} for an explanation for why more
rules improve stability).
The
rules are assumed to be constructed independently of the data used by
VarPro, and therefore without loss of generality, it will be assumed that
all rules are deterministic.
For each $n$, let $\z_{n,1},\ldots,\z_{n,K_n}$ denote these
rules.  Notice that the number of rules $K_n$ can vary with $n$ and
also that the rules themselves are allowed to change with $n$.  For a
given rule $\z$, define
$$
\that_n(\z)=m_n(\z)^{-1}\sum_{i=1}^n  g(Y_i)I\{\X_i\in R(\z)\},
\hskip10pt
m_n(\z)=\sum_{i=1}^n I\{\X_i\in R(\z)\},
$$
which is a slightly more compact way of writing $\that_n$
than~\mref{varpro.estimator} (notice that $m_n(\z)$ equals the sample size of
a rule $\z$).
In a likewise fashion, we can define
$\that_n(\z^S)=m_n(\z^S)^{-1}\sum_{i=1}^n  g(Y_i)I\{\X_i\in
R(\z^S)\}$.

For notational ease,
let $m_{n,k}=m_n(\z_{n,k})$, $m_{n,k}^S=m_n(\z_{n,k}^S)$
and $R_{n,k}=R(\z_{n,k})$, $R_{n,k}^S=R(\z_{n,k}^S)$.
The VarPro importance score for $S$ is defined as the weighted averaged difference
\EqArray
\D_n(S)
&=& \sum_{k=1}^{K_n} W_{n,k}|\that_{n}(\z_{n,k}^S)-
\that_{n}(\z_{n,k})|
\label{varpro.importance}\\
&=& \sum_{k=1}^{K_n} W_{n,k}\Bigg|
\frac{1}{m_{n,k}^S}\sum_{i=1}^n  g(Y_i)I\{\X_i\in R_{n,k}^S\}
-\frac{1}{m_{n,k}}\sum_{i=1}^n  g(Y_i)I\{\X_i\in R_{n,k}\}
\Bigg|
\nonumber
\EndEqArray
where $0\le W_{n,k}\le 1$ are weights (deterministic or random)
such that $\sum_{k=1}^{K_n}W_{n,k}=1$. In the following
sections, we study the asymptotic properties
of~\mref{varpro.importance}, breaking this up into the case of noise
and signal variables.

\subsection{Consistency for noise variables}

The following assumptions will be used.
Our first assumption requires that:
\Enumerate
\setlength\itemsep{5pt}
\item[(A1)]
$\E[g(Y)^2]<\infty$ and $\E[\psi(\X)^2]<\infty$.
\EndEnumerate
We also require $\psi$ satisfies a smoothness property and that
the features space is connected.
\Enumerate
\setlength\itemsep{5pt}
\item[(A2)]
$\psi$ is continuous and differentiable over the connected space $\xx\subseteq\RR^p$
and possesses a gradient $\f=\grad\psi:\xx\rightarrow\RR^p$ 
satisfying the Lipschitz condition
\Eq
|\fs(\x_1)-\fs(\x_2)|\le C_0 |\xss_1-\xss_2|,
\qt{for all }\x_1,\x_2\in\xx,
\label{lipschitz.assume}
\EndEq
for some constant $C_0<\infty$
where $\f^{(S)}$ denotes the subvector of $\f$ with coordinates in
$S$ (note that the Lipschitz condition
only applies to $\fs$ as $\f$ is zero over the
coordinates from $\nn$).
\EndEnumerate
We require that regions shrink to zero uniformly over the signal
features (the rate is unspecified):
\Enumerate
\setlength\itemsep{5pt}
\item[(A3)]
For $k=1,\ldots,K_n$,  $\PP\{\X\in R_{n,k}\}>0$ and $\diam_{\ss}(R_{n,k})\le r_n$ for
some sequence $r_n\rightarrow 0$ where
$\diam_{\ss}(R_{n,k})=\sup_{\{\x_1,\x_2\in
  R_{n,k}\}}||\xss_1-\xss_2||_2$.
\EndEnumerate
The last condition pertains to the weights:
\Enumerate
\setlength\itemsep{5pt}
\item[(A4)]
For each $n$, there exists $\x_{n,k}\in R_{n,k}$ for $k=1,\ldots,K_n$
such that
\Eq
\sum_{k=1} W_{n,k} |\psi(\x_{n,k})| \le C_1,\qt{}
\sum_{k=1} W_{n,k} ||\f(\x_{n,k})||_2 \le C_2,
\label{weight.integrability}
\EndEq
for some constants $C_1,C_2<\infty$.
\EndEnumerate
Assumption (A4) stipulates a trade off between the weights and the
behavior of $\psi$ and $\f$ over the region of the feature space
identified by a rule.  Condition~\mref{weight.integrability} is
satisfied if $\psi$ and $\f$ are bounded.  The condition
also holds if all rules are constructed so their regions are
contained within some closed bounded subspace in $\xx$ defined by the
signal coordinates (because $\psi$ and $\f$ will be bounded due to
continuity).

The following theorem shows that under these assumptions,
VarPro is consistent for noise variables.

\begin{theorem}\label{varpro.noise.theorem}
Assume that (A1), (A2), (A3) and (A4) hold.  If
$K_n\le O(\log n)$ and $m_{n,k}\ge m_n=n^{1/2}\g_n$ where
$\g_n\uparrow\infty$ at a rate faster than $\log n$,
then $\D_n(S) \prob 0$ if $S\subseteq\nn$.
\end{theorem}

Going back to our assumptions, $\that_{n}(\z_{n,k}^S)$ and
$\that_{n}(\z_{n,k})$ from~\mref{varpro.importance} are averages of
independent variables.  Seen in this light, (A1) is really just a
second moment condition needed to ensure that they 
converge.  Assumptions (A2) and (A4) are smoothness and boundedness
conditions for $\psi$ and its derivative.  Roughly speaking, since
$m_n\Gt n^{1/2}\rightarrow\infty$, then due to the large sample
properties of averages, $\that_{n}(\z_{n,k}^S)- \that_{n}(\z_{n,k})$
converges to $\E(\psi(\X)|\X\in R_{n,k}^S)-\E(\psi(\X)|\X\in
R_{n,k})$.  For this to provide useful information about a variable's
importance, it is necessary for $\psi$ to have certain nice
properties.  Regarding the number of rules $K_n$ used by the
estimator, we want to use as many rules as possible to improve
stability, but there is a trade-off because we also have to maintain
uniform convergence, which holds when
$K_n\le O(\log n)$.

Assumption (A3) and the rate condition $m_n$ are the only conditions
that specifically apply to a region.  (A3) requires that the diameter
of a region shrinks to zero uniformly over the signal variables while
$m_n$ places a lower bound on the sample size of a region which must
increase to $\infty$.  While these two conditions are
diametrically opposed and therefore might seem unusual, they are 
fairly standard assumptions used 
in the asymptotic analysis of trees.  For
tree consistency, a typical requirement is that the diameter of a
terminal node converges to 0 and the number of its points converges to
$\infty$ in probability.  See Theorem 6.1
from~\cite{devroye2013probabilistic} and Theorem 4.2
from~\cite{gyorfi2002distribution}.

In fact, we directly show that conditions (A3) and the rate condition
$m_n$ hold for a random tree construction. Assume that $\xx =
[0,1]^p$. Consider a tree construction where at the start of split $k
\ge 1$, each of the $k$ leaves of the tree is a $p$-dimensional
rectangle (when $k=1$, the one leaf equals the root node of the tree,
$[0,1]^p$). Among these $k$ leaves (rectangles), one is selected at
random and its longest side is split at a random point; yielding, at
the end of step $k$, two new rectangles of reduced volume from the
original rectangle. The tree construction is repeated for a total of
$K_n$ splits, yielding $K_n+1$ branches which are the rules.

If $K_n=\log n$, then it can be shown that the following holds in
probability~\citep[see the proof of Theorem 2 of][]{biau2008consistency}:
\Enumerate
\setlength\itemsep{5pt}
\item[(i)]
The number of data points in a rectangle is greater than
$m_n=n^{\d}\log n$ for any $1/2<\d<1$.
\item[(ii)]
The mean length of the largest side of a
rectangle is less than $\E\[(3/4)^{T_n}\]$ where $T_n\prob\infty$ does not depend
on the specific rectangle;
hence the diameter of each rectangle converges
uniformly to zero.
\EndEnumerate
Therefore, (ii) shows (A3) holds.  Also, 
(i) shows that $m_n\Gt n^{1/2}$ achieves the lower
bound required by~\cref{varpro.noise.theorem}.  This is an interesting consequence of the random
construction and allows the bound to be achieved without
supervision.   This can be seen informally from the fact that
$K_n=O(\log n)$ which  naturally forces data to
pile up in the leaves and if 
tree node sizes are approximately evenly distributed,
$m_n\asymp n/\log n$.

\subsection{Limiting behavior for signal variables}

For the analysis of signal variables, we assume that
regions are rectangles.  This is made for
technical reasons to simplify arguments but does not limit 
applications of VarPro.  Thus, 
the region for a rule $\z$ can be written as $R(\z)=\bigtimes_{j=1}^p I_{j}$
where $I_{j}\subseteq\RR$ are real-valued intervals.  For notational
simplicity we assume the first $|\ss|$ coordinates are signal
variables and the remaining $|\nn|$ coordinates are noise variables.
Therefore,  we can write $R(\z)=A\bigtimes B$ where $A=\bigtimes_{j=1}^{d}
I_{j}$, $B=\bigtimes_{j=d+1}^p I_{j}$ and $d=|\ss|\le p$ is the number
of signal variables.

\begin{theorem}\label{varpro.signal.theorem}
Assume that the region for each rule is a rectangle contained within
the connected space $\xx\subseteq\RR^p$, then
under the same conditions as Theorem~\mref{varpro.noise.theorem},
if $S=\{s\}$ is a signal variable,
$$
\D_n(s) =
\Big(1+o_p(1)\Big)\sum_{k=1}^{K_n} W_{n,k}
\Big|\E\[\big(\psi_{n,k}^s(\Xs{s})-\psi_{n,k}^s(x_{n,k}^{(s)})\big)
\,\big|\,\X\in R_{n,k}^s\]\Big|
+o_p(1),
$$
where 
$\psi_{n,k}^s(z)=\psi(x_{n,k}^{(1)},\ldots,x_{n,k}^{(s-1)},z,x_{n,k}^{(s+1)},\ldots,x_{n,k}^d)$
and $\x_{n,k}=(x_{n,k}^{(1)},\ldots,x_{n,k}^{(p)})'\in R_{n,k}$ is
a fixed point in each rectangle as defined in (A4).
\end{theorem}

\cref{varpro.signal.theorem} shows that the limit of the VarPro estimator is not
necessarily zero as in the noise case.
The limit is complicated as we would expect since it will depend on
the size of the signal as well as the feature distribution, however
to help understand what this limit might be, consider the case
when $\psi$ is an additive function, $\psi(\x)=\sum_{j=1}^d
\phi_j(\xs{j})$.  Then by definition $\psi_{n,k}^s(x)=\sum_{j\in\ss\setminus s}
\phi_j(x_{n,k}^{(j)}) + \phi_s(x)$ and
\Eq
\E\[\(\psi_{n,k}^s(\Xs{s})-\psi_{n,k}^s(x_{n,k}^{(s)})\)\,\big|\,\X\in R_{n,k}^s\]
= \E\[\(\phi_s(\Xs{s})-\phi_s(x_{n,k}^{(s)})\)\,\big|\,\X\in R_{n,k}^s\].
\label{additive.signal}
\EndEq
This is the average difference between $\phi_s(\Xs{s})$ for a random
$\X\in R_{n,k}^s$ compared with $\phi_s(x_{n,k}^{(s)})$ for a fixed
point $\x_{n,k}\in R_{n,k}$.  Because $\Xs{s}$ is unrestricted, and
can take any value in the $s$ coordinate direction of $\xx$,
$\phi_s(\Xs{s})-\phi_s(x_{n,k}^{(s)})$ should be non-zero on average.
Also, even if~\mref{additive.signal} equals zero by chance, keep in
mind this applies to each rectangle $R_{n,1},\ldots,R_{n,K_n}$, thus
we can expect an average nonzero effect when summing over all rules.
This also explains why it is better to use many rules than just one
rule.  In fact,~\cref{varpro.signal.theorem} allows for up to $O(\log
n)$ rules.

\section{Empirical results}

Here we study the performance of VarPro in regression and
classification problems. \cref{A:varpro.guided} describes the
computational procedure used for our analysis.
In line 2, the data of size $N$ is split
randomly into two parts. In line 3, the first dataset of size $N - n$
is used for rule generation. In line 4, the second dataset of size $n$
is used to obtain the importance score using the rules obtained in
line 3. Data split proportions used are $0.632N$ for the rule
generation step and $n = (1 - 0.632)N$ for calculating the importance
score. This type of data split is fairly common in machine
learning. For example, in permutation importance of random forests,
each tree is constructed from a bootstrap sample of approximately
$63.2\% \times N$ and the remaining $36.8\% \times N$ (the so-called
out-of-bag data) is put aside for calculating permutation importance.

\vskip10pt
\begin{algorithm}[phtb]
  \centering
\caption{\em\,\, VarPro for Model-Independent Variable Selection}\label{A:varpro.guided}
\begin{algorithmic}[1]
  \FOR {$b=1,\ldots,B>0$}
  \STATE Split the data $D$ of size $N$ randomly into separate parts
  $D=\Drg\cup \Dvp$ where $\Drg$ is of
  of size $N-n$ and $\Dvp$ is of size
  $n=(1-\a)N$ where $\a=.632$.
  \STATE Use $\Drg$ for the rule generation step, yielding rules $R_{n,1},\ldots,R_{n,K_n}$.
  \STATE Using $\Dvp$, calculate the VarPro importance~\mref{varpro.importance} for variable $\Xs{s}$ 
      for $s=1,\ldots,p$ using rules
      $R_{n,1},\ldots,R_{n,K_n}$ generated in the previous step.  When calculating~\mref{varpro.importance} 
      use weights $W_{n,k}=m_{n,k}/\sum_{k=1}^{K_n} m_{n,k}$. 
        Let $\D_{n}^b(s)$ denote the importance value for $s=1,\ldots,p$.
  \ENDFOR
  \STATE
  Calculate the sample average $\overline{\D}_{n}(s)$ and sample variance
  $\Var_n(s)$ of $\{\D_{n}^1(s),
  \ldots, \D_{n}^B(s)\}$.  Define
  \Eq
  I_{n,B}(\Xs{s}) = 
   \frac{\overline{\D}_{n}(s)}{\sqrt{\Var_n(s)}} ,\hskip10pt
  s=1,\ldots,p.
  \label{bootstrap.standvimp}
  \EndEq
  Call~\mref{bootstrap.standvimp} the VarPro standardized importance
  of $\Xs{s}$.
\end{algorithmic}
\end{algorithm}

We use trees for the rule generating procedure (line 3). Our trees are
constructed using a guided tree-splitting strategy. As in random
forests, the tree is grown using random feature selection where the
split for an internal node is determined from a randomly chosen subset
of features. However, instead of a uniform distribution, features are
selected with probability proportional to a pre-calculated
split-weight. This is done to encourage the selection of signal
features in the tree construction in accordance with assumption
(A3). The split-weights are obtained prior to growing the tree by
taking the standardized regression coefficients from a lasso fit and
adding these to the variable's split frequency obtained from a
previously constructed forest of shallow trees. The rationale for this
approach is that combining lasso with trees borrows the strengths of
parametric and nonparametric modeling. After constructing the tree,
$K_n$ branches are randomly chosen, yielding the rules $R_{n,1},
\ldots, R_{n,K_n}$ of line 3.

Finally, for the purpose of
reducing variance and producing a standardized importance value, the
entire procedure is repeated $B$ times and the importance scores are
standardized (line 6).  Variable
$\Xs{s}$ is selected if~\mref{bootstrap.standvimp} exceeds a cutoff
value $Z_0$.  This cutoff can be pre-chosen (for example $Z_0=2$) or
selected by out-of-sample validation.  Examples using both strategies
will be presented.

\subsection{Regression}

Regression performance of VarPro was tested on 20 synthetic datasets
with different $N$ and $p$ (see~\cref{appE}) covering linear and
nonlinear models as well as models that switched between between
linear and nonlinear.  Features varied from uncorrelated to
correlated, with the latter retaining the same distribution but
adjusted to a 0.9 correlation via a copula, except for specific cases
like {\it lm} and {\it lmi2} where features were correlated within
blocks of size 5 (details are provided in~\cref{appE}).

Each VarPro run used $B=500$ with $K_n=75$ rules extracted.
Comparison methods included permutation importance~\citep{Breiman2002}
(referred to as Breiman-Cutler variable importance, abbreviated as
BC-VIMP), generalized boosted regression modeling (GBM) with trees;
i.e.~gradient boosted trees~\citep{Friedman2001},
knockoffs~\citep{candes2018panning},
lasso~\citep{tibshirani1996regression}, mean decrease impurity
(MDI)~\citep{louppe2013understanding},
wrappers~\citep{kohavi1997wrappers} and the Relief
algorithm~\citep{kira1992feature} via its adaptation
RReliefF~\citep{robnik1997adaptation}.

R-packages used for the analysis included:
\rfsrc~\citep{rfsrc},
\gbm~\citep{gbmr},
\knockoff~\citep{knockoffr},
\glmnet~\citep{friedman2010regularization},
\ranger~\citep{ranger},
{\ttfamily mlr}~\citep{mlr} (using a sequential forward
search engine)
and 
\core~\citep{robnik1997adaptation}.
Two types of knockoff test statistics were calculated from differences
using: (1) lasso coefficient estimates and (2) random forest impurity
compared to their knockoff counterparts. Two types of wrappers were
used: (1) $k$-nearest neighbors (wrapper KNN) and (2) boosted gradient
trees (wrapper GBM). The lasso regularization parameter, the number of
boosting iterations for GBM and wrapper GBM and the number of nearest
neighbors $k$ for wrapper KNN were determined by 10-fold
cross-validation.

Feature selection performance was assessed using the area under the
precision-recall curve (AUC-PR) and the geometric mean (gmean) of TPR
(true positive rate for selecting signal variables; i.e.~the
sensitivity) and TNR (true negative rate for selecting noise
variables; i.e.~the specificity). These metrics were selected since
many of the simulations had far more noise variables than signal
variables, presenting an imbalanced classification problem. AUC-PR
evaluates the trade-off between TPR (recall) and positive predictive
value (precision) without being affected by the imbalance ratio since
the recall and precision are evaluated by selection thresholds varied
over all possible values, making it suitable for imbalanced
datasets. Gmean provides a balanced measure of performance across both
the majority and minority classes, and is therefore is also appropriate
for class imbalanced situations~\citep{kubat1997learning}.

To calculate AUC-PR, which does not require a threshold value, a
method's output was converted to a continuous score.  Scores used
were: For BC-VIMP, permutation importance (a value that can be both
positive and negative); for GBM, the relative influence (a
non-negative value); for knockoffs, the absolute value of the knockoff
test statistic; for lasso, the absolute value of the coefficient
estimates; for MDI the impurity score; for ReliefF, the attribute
evaluation score; for wrappers, this was a zero-one value reflecting if
a variable was selected; for VarPro, the standardized importance
value~\mref{bootstrap.standvimp} (a non-negative value).

For gmean, a zero threshold was uniformly applied. GBM, lasso, MDI,
ReliefF and
wrappers used  the same score value as their AUC-PR calculations. For
BC-VIMP, negative importance values were converted to zero. For
knockoffs, the knockoff test statistic was set to zero for variables
screened under a target FDR value of 0.1. For VarPro, standardized
importance was set to zero for values less than an adaptive cutoff
value $Z_0$ selected using an out-of-sample approach. In this strategy, a
grid of cutoff values is used. After ranking features in descending
order of their VarPro importance, a random forest is fit to those
features whose VarPro importance value is within a given cutoff value,
and the out-of-bag error rate for the forest is stored. The optimized
$Z_0$ value is defined as the cutoff with smallest out-of-bag error.

\begin{figure}[phtb]
  \vskip10pt
  \centering
  \resizebox{6in}{!}{\includegraphics[page=2]{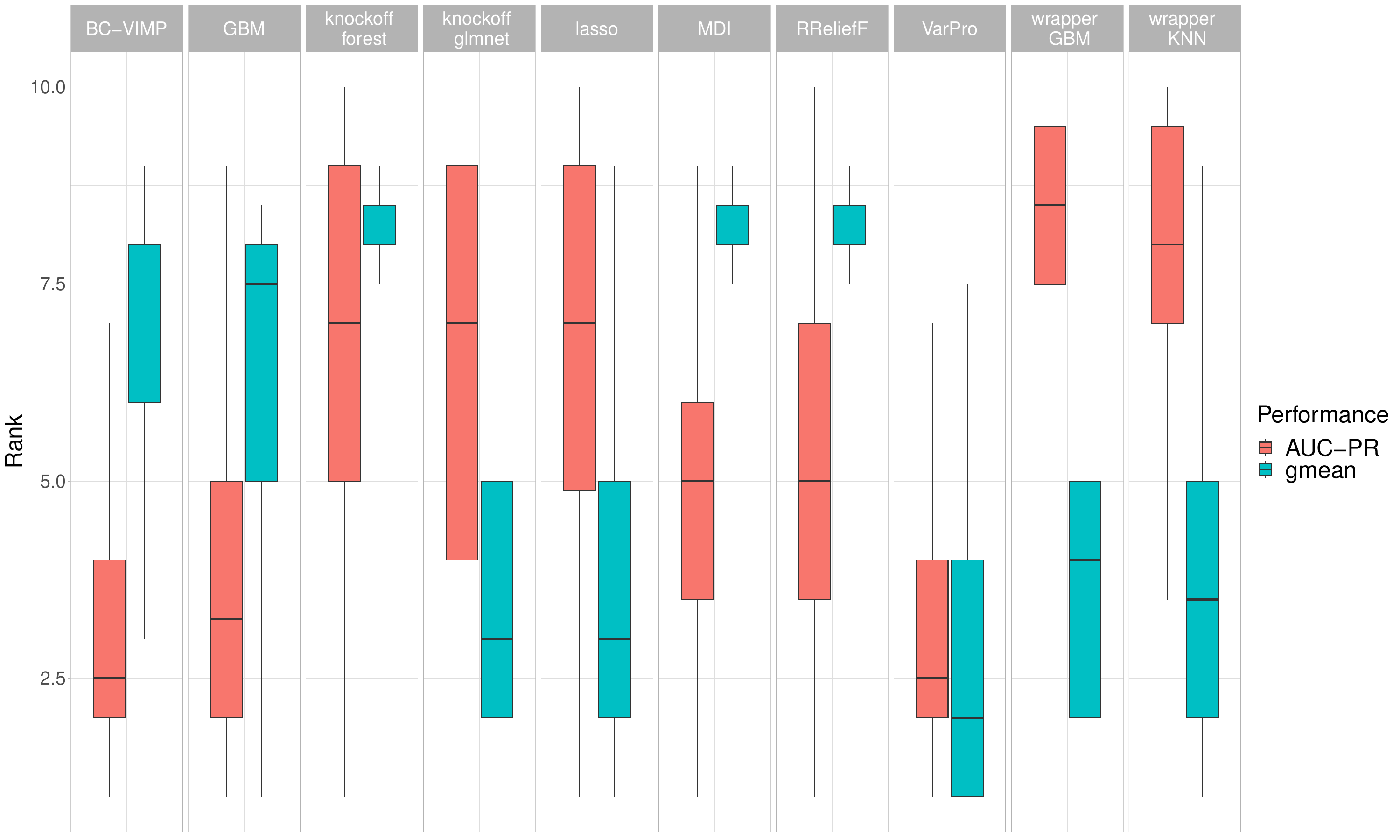}}
\caption{\small Rank of each procedure for correlated
  regression simulation experiments (lower indicates better performance).}
\label{benchmark.rank.correlated}
\end{figure}

Each experiment was run 100 times independently. Results for the
uncorrelated and correlated feature experiments are shown in
\cref{benchmark.uncorrelated} and~\cref{benchmark.correlated}
(\cref{appE}). We will focus on the more challenging correlated
simulations. This is summarized in~\cref{benchmark.rank.correlated},
which shows the rank of each procedure for the correlated
setting. Considering~\cref{benchmark.correlated}
and~\cref{benchmark.rank.correlated}, we can observe that BC-VIMP and
GBM generally have similar behavior. Both achieve high AUC-PR values
but both also have poor gmean performance. This is because while they
rank variables reasonably well, which yields high AUC-PR, both methods
tend to overfit and they are not able to threshold variables
effectively, thus leading to a low gmean metric. The lasso and glmnet
knockoffs show more balanced gmean results but face issues with
complex models like \textit{inx1}, \textit{inx2}, and \textit{inx3},
affecting their consistent performance across experiments for both
metrics. Forest knockoffs generally underperform compared to
glmnet. The KNN and GBM wrappers perform similarly and are not as
effective as lasso and glmnet knockoffs. RReliefF and MDI both perform
poorly in terms of gmean, with MDI slightly outperforming RReliefF in
terms of AUC-PR.  VarPro stands out with the best overall performance,
achieving high AUC-PR for accurate variable ranking and high gmean for
effective signal and noise differentiation. To test if VarPro's
performance is statistically superior, we applied a paired one-sided
Wilcoxon signed rank test with a Bonferroni correction for multiple
comparisons. The only method equivalent to VarPro is BC-VIMP in terms
of AUC-PR (adjusted $p$-value <.00001 for all other methods), and no
method is equivalent or is able to outperform VarPro in terms of gmean
(adjusted $p$-value <.00001). Therefore, this demonstrates VarPro has
robust overall performance.

\subsection{Classification}

The following synthetic example was used to compare VarPro to random
forest permutation importance.  The simulation model used was a
multiclass model with $L=3$ specified according to:
\Array
y &=& \argmax_{l} \{\psi_l(\x)\},\hskip10pt\text{where }
\psi_l(\x)
= \dfrac{\phi_{l}(\x)}{\sum_{l=1}^L \phi_{l}(\x)} ,\hskip3pt
\phi_{l}(\x)=\exp\(\sum_{j=1}^p \beta_{j,l}\xs{j}\)\\
&&\beta_{j,1}=1\,\,\text{for } j=1,2,3,\,  \text{otherwise} \,\beta_{j,1}=0\\
&&\beta_{j,2}=1\,\,\text{for } j=4,5,6,\,  \text{otherwise} \,\beta_{j,2}=0\\
&&\beta_{j,3}=1\,\,\text{for } j=7,8,9,\, \text{otherwise}\, \beta_{j,3}=0.
\EndArray
Here $\Xs{1},\ldots,\Xs{9}$ are signal variables for all
three classes due to the constraint $\sum_l\psi_l=1$.  However,
coordinates 1,2,3 are especially informative for class 1, coordinates
4,5,6 for class 2, and coordinates 7,8,9 for class 3.  The features
were drawn from a multivariate normal with marginals $\Xs{j}\sim N(0,
1)$ such that all coordinates were independent of one another, except
for pairs $(\Xs{3},\Xs{10})$, $(\Xs{6},\Xs{15})$, $(\Xs{9},\Xs{20})$
which were correlated with correlation $\rho=.9$.  Simulations used
$N=2000$ and $p=20$.

\begin{figure}[phtb] 
\centering
\resizebox{6.25in}{!}{\includegraphics[page=1]{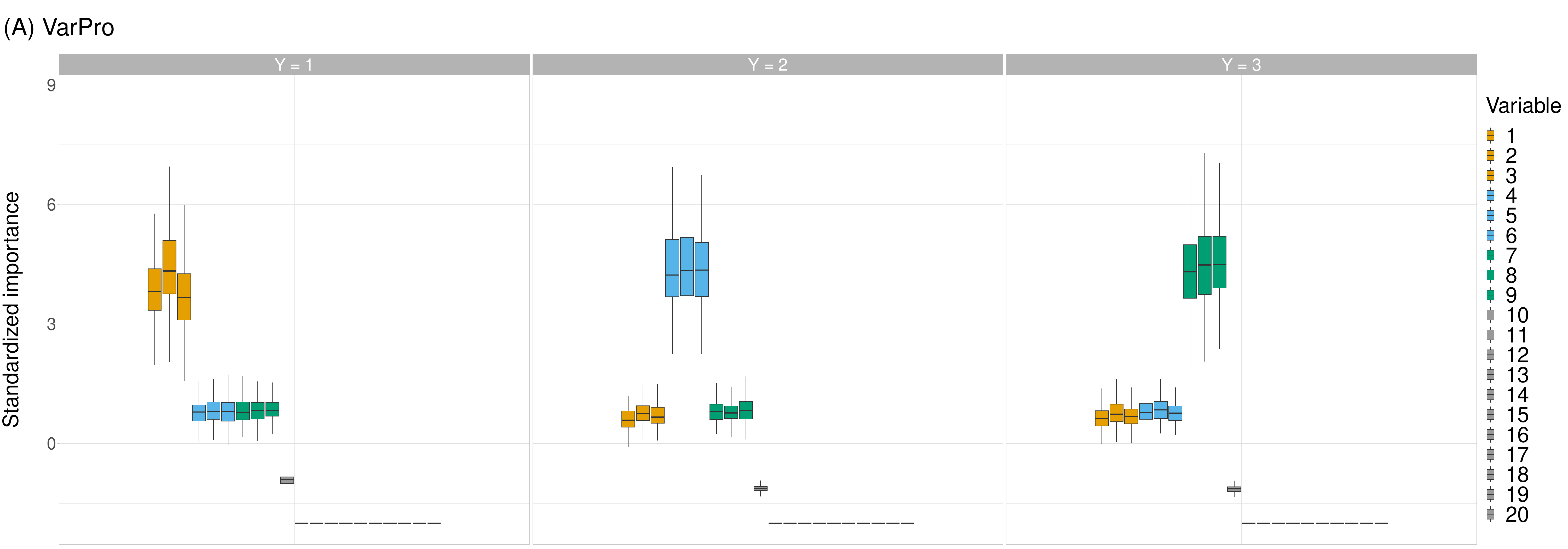}}
\vskip30pt
\resizebox{6.25in}{!}{\includegraphics[page=2]{multiclassification.pdf}}
\vskip10pt
\caption{\small Multiclass experiment where variables 1--3 are
  most informative for class 1, variables 4--6 for class 2 and variables
  7--9 for class 3; variables 10--20 are noise variables. Variables
  3 and 10, 6 and 15, 9 and 20 are strongly correlated: thus there is
  correlation between signal and noise features.  (A) VarPro
  importance correctly
  identifies the group structure and is not influenced
  by correlation. (B) BC-VIMP from random
  forests is influenced by correlation that degrades its
  performance.}
\label{multiclass}
\end{figure}

Standardized VarPro importance values were subtracted by the constant
$Z_0=2$, yielding a standardized importance value for each variable
$\Xs{s}$, $s\in\{1,\ldots,p\}$, calibrated to zero for each class.
The results from 250 independent runs are given in \cref{multiclass}.
VarPro's results shown in the top panel are excellent which is in
contrast to the bottom panel displaying permutation importance
(BC-VIMP).  The latter shows clear issues in separating noise from
signal.  The poor performance of BC-VIMP is due to the correlation
between the signal variables $\Xs{3},\Xs{6},\Xs{9}$ and noise
variables $\Xs{10},\Xs{15},\Xs{20}$ which causes permutation
importance for signal to be degraded while artificially inflating
noise importance values.  In each class label instance, BC-VIMP incorrectly
identifies 2 or 3 noise variables as signal.  This differs from VarPro
where signal variable importance is not degraded and noise 
importance scores are substantially smaller and all are non-significant. The group
structure is also clear (for example features $\Xs{1},\Xs{2},\Xs{3}$
are highly informative for class 1) and it is evident that each
conditional probability depends on $\Xs{1},\ldots,\Xs{9}$.

\vskip10pt

\subsubsection{Low sample size, high-dimensional microarray
  classification experiments}

In our next example we run a benchmark experiment using a collection
of 22 small sample size, high-dimensional microarray datasets.  The
sample sizes vary from $N=62$ to $N=248$ and the dimensions vary from
$p=456$ to $p=54613$.  In each example the outcome is a class
label where the number of classes $L$ vary from $L=2$ to $L=10$.
Most of the datasets are related to cancer.  The data is available
from the R-package \texttt{datamicroarray} and is described in more
detail there.

For each dataset, the original class labels were replaced by a
synthetically generated set of class labels.  A parametric lasso model
was fit using the original $Y$ which was regressed using multinomial
regression against $\X$ equal to the gene expression values.  Three
different synthetic $Y$'s were created.  In the first, $Y$ was the
predicted class from the preliminary lasso model. In
the second, the expression values were squared and then applied to the
original estimated coefficients to obtain $Y$.  The third $Y$ was
similar to the second simulation except that the top three features
were replaced with their pairwise interactions.  Thus we have three
simulation models: (A) linear, (B) quadratic, (C)
quadratic-interaction.  Note that the simulated datasets use the
original $\X$, only the $Y$ is replaced with the synthetic $Y$ and in
each simulation we know which features are signal and which features
are noise.

VarPro feature selection used an optimized $Z_0$ cutoff using
out-of-sampling similar to the regression benchmark experiments. For
comparison procedures, several well-known methods suitable for dealing
with high-dimensional data were used: (1) BC-VIMP using the R-package
\texttt{rfsrc}~\citep{rfsrc}. (2) MCFS (Monte Carlo Feature Selection) using
the R-package \texttt{rmcfs}~\citep{draminski2018rmcfs}. MCFS
calculations were performed using the function \texttt{mcfs} where the
cutoff value for thresholding was determined by permutation. As
recommended by the \texttt{rmcfs} package, we used a large number of
permutations and ran the MCFS algorithm for each permutation, which
was used by the package to determine the permutation threshold. (3)
SES (Statistically Equivalent Signature) which uses constraint-based
learning of Bayesian networks for feature selection. Calculations used
the \texttt{SES} function from the \texttt{MXM}
R-package~\citep{mxm}. (4) Boruta which is a feature filtering
algorithm that creates artificial data in the form of ``shadow
variables'' which act as a reference for filtering noise
variables~\citep{boruta}. Boruta computations were implemented using
the \texttt{Boruta} R-package using random forests for the training
classifier. (5) Recursive Feature Elimination (RFE) which is a
wrapper feature selection procedure~\citep{guyon2002gene}.
Calculations used the \texttt{rfe} function from the \texttt{caret}
R-package~\citep{caret}. Random forests were used as the RFE wrapper
as recommended by~\cite{chen2020selecting}.

The simulations were run 100 times independently for each microarray
dataset.  In order to introduce variability across runs, rather than
using all $p$ variables, a random subset of two-thirds of the noise
variables were selected where a noise variable was defined as a gene
expression feature not selected in the preliminary lasso model.
Feature selection performance was assessed by geometric mean (gmean),
precision (1 minus false discovery rate) and accuracy (1 minus
misclassification error). Values were averaged over the 100 runs.

\begin{figure}[phtb]
  \centering
(A) Linear\hspace*{4in}\hfill\\[10pt]
\resizebox{5.5in}{!}{\includegraphics[page=1]{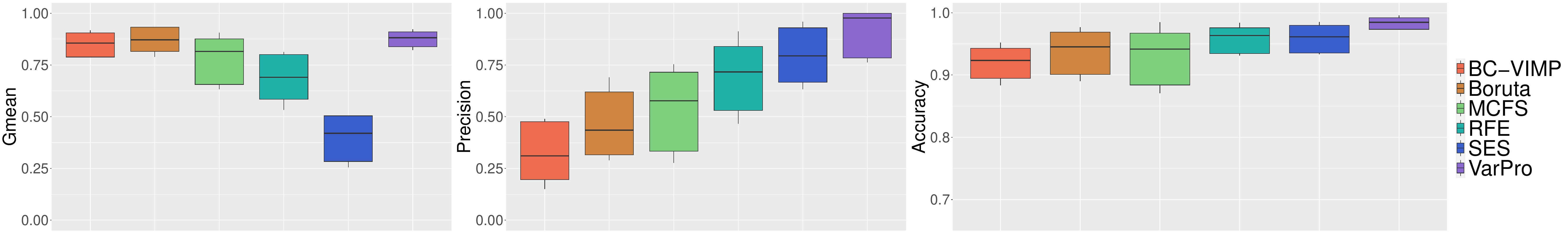}}
\vskip15pt
(B) Quadratic\hspace*{4in}\hfill\\[10pt]
\resizebox{5.5in}{!}{\includegraphics[page=2]{microarray.pdf}}
\vskip15pt
(C) Quadratic-Interaction\hspace*{4in}\hfill\\[10pt]
\resizebox{5.5in}{!}{\includegraphics[page=3]{microarray.pdf}}
\vskip10pt
\caption{\small Feature selection performance over
  high-dimensional microarray classification datasets.  Three simulation
  models were used for creating synthetic $Y$ classification labels:
  (A) Linear; (B) Quadratic; (C) Quadratic-Interaction.}
\label{microarray.performance}
\end{figure}

The results are recorded in~\cref{microarray.performance}.
Going from scenarios (A) to (C), performance for all
methods decreases as the complexity of the simulation increases as is
to be expected.  However, at the same time, the rank of each method
and their various characteristics are maintained across simulations.
BC-VIMP is always one of the best procedures in terms of gmean whereas
it is always one of the worst for precision and accuracy.  SES is
always the worst in terms of gmean performance, but it is always among
the top performers for precision and accuracy.  VarPro, on the other
hand, does well in all three metrics and has better overall
performance than any other method (adjusted $p$-value < 0.05 for all
experiments).  These results are consistent with our benchmark
regression experiments that showed VarPro achieves a balanced
performance.

\subsubsection{DNA methylation subtypes for adult glioma multiclassification}

In our next example we reanalyze a subset
of data used by~\cite{ceccarelli2016molecular} for studying
adult diffuse gliomas.  The study used DNA
methylation data obtained from CpG probes for molecular profiling.  A
total of $p=1206$ probes collected over $N=880$ human tissues was used
in their analysis.

An important finding in~\cite{ceccarelli2016molecular} was that DNA
methylation was effective in separating glioma subtypes. To informally
validate these results, we use the clusters developed in the study as
outcomes in a multiclass analysis. There were $L=7$ labels:
Classic-like, Codel, G-CIMP-high, G-CIMP-low, LGm6-GBM,
Mesenchymal-like, and PA-like. For an added challenge, clinical data
and molecular data available for the samples were also included as
features for the VarPro analysis. In total, $p=1241$ variables were
used. The data is available from the R-package
\texttt{TCGAbiolinksGUI.data}~\citep{TCGAbiolinksGUI.data}.

\begin{figure}[phtb]
\vskip-10pt
\centering
\resizebox{5.0in}{!}{\includegraphics[page=1]{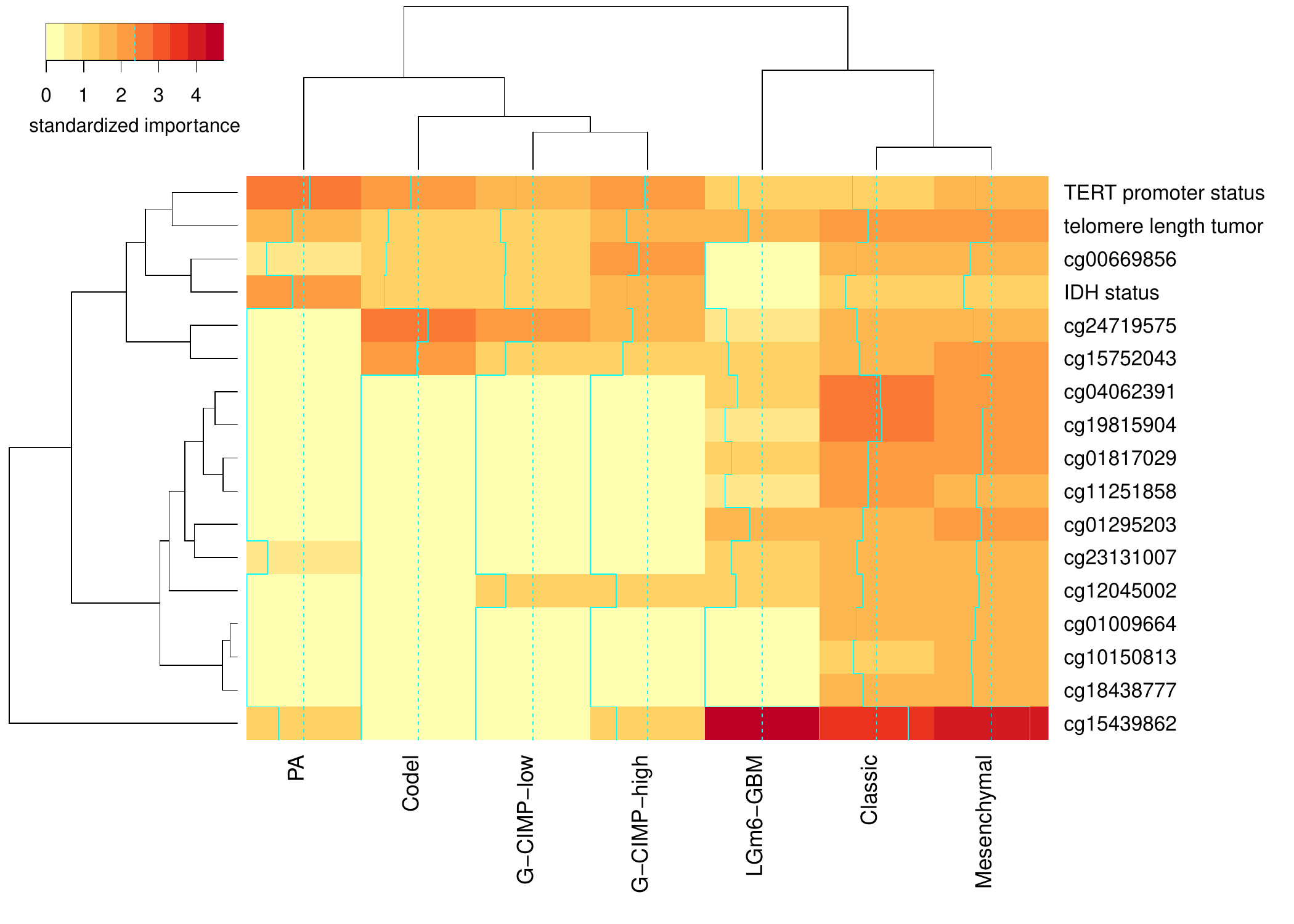}}
\vskip-10pt
\caption{\small Heatmap of standardized importance from VarPro multiclass
  analysis of adult glioma DNA methylation data ($N=880$, $p=1241$).} 
\label{tcga}
\vskip20pt
\resizebox{6.0in}{!}{\includegraphics[page=1]{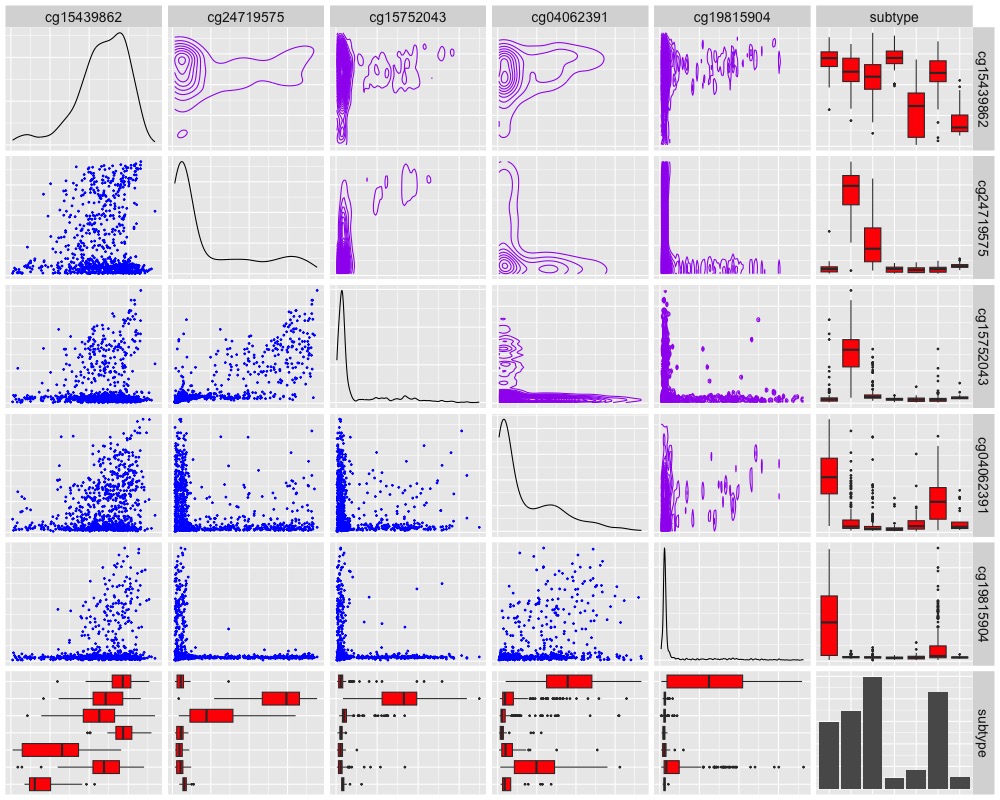}}
\caption{\small DNA methylation probe values and glioma
  subtypes.}
\label{tcga.probes}
\end{figure}

\cref{tcga} displays the VarPro standardized importance values for the
selected features. IDH status, telomere length, and TERT promoter
status, which were three of the added molecular variables, are seen to
be informative. \cite{ceccarelli2016molecular} found that IDH status
was a major determinant of the molecular profile of diffuse glioma, so
these results are in agreement with the original study. The VarPro
analysis also identified DNA methylation probes with significant
importance values. This is interesting and useful since it identifies
probes that potentially provide additional power to separate subtypes
beyond the molecular variables mentioned.

To explore the ability of probes to separate
subtypes,~\cref{tcga.probes} shows some of the top DNA methylation
probes versus glioma subtypes. Data distribution for each variable and
frequency distribution of cluster subtype is given along the
diagonal. Lower subdiagonal blocks in blue are scatter plots of probe
values in pairs; upper subdiagonal in purple are density contours for
the pairs. Boxplots in red show DNA methylation probe values for each
subtype. For example, the top right boxplot is cg15439862, which shows
particularly low values for LGm6-GBM and Mesenchymal compared to other
subtypes, indicating this probe's effectiveness in distinguishing
these two subtypes from the rest. Similar patterns are observed with
other probes, demonstrating their ability to separate subtypes. These
results support the conclusions of the original study and highlight
VarPro’s effectiveness in addressing high-dimensional bioinformatic
challenges.

\section{Modified VarPro using external estimators: time to
  event data}

Now we discuss the use of external estimators, which is a way to
extend the VarPro method from being model-independent to
model-specific. Recall that VarPro estimates $\psi(\X) = \E(g(Y) \mid
\X)$ by using a sample average of $g(Y)$ calculated locally within a
region defined by a rule. For a rule $\z_{n,k}$ and its region
$R_{n,k} = R(\z_{n,k})$ with sample size $m_{n,k}$, VarPro makes use
of the estimator~\mref{varpro.estimator}
$$
\that_n(\z_{n,k})
=\frac{1}{m_{n,k}}\sum_{i=1}^n g(Y_i)I{\{\X_i\in R(\z_{n,k})\}}.
$$
However, depending on the
complexity of the problem, this may not always be suitable 
as it could be inefficient or difficult to estimate $\psi$
locally this way.

The strategy to overcome this is to use an external estimator
building on the ideas discussed in Section 2.
For each $n$, let $\psi_n:\xx\rightarrow\RR$ be an external estimator
for $\psi$.
The modified VarPro estimator is defined as follows.
For each rule $\z_{n,k}$, the modified procedure replaces
$\that_n(\z_{n,k})$ with
$$
\ts_n(\z_{n,k})=
\frac{1}{m_{n,k}}
\sum_{i=1}^n\psi_n(\X_i)I{\{\X_i\in R_{n,k}\}}.
$$
Likewise, the release rule $\that_n(\z_{n,k}^S)$, which releases the
rule $\zeta_{n,k}$ along the coordinates of a set
$S\subset\{1,\ldots,p\}$, is replaced with
$$
\ts_n(\z_{n,k}^S)=\frac{1}{m_{n,k}^S}\sum_{i=1}^n
\psi_n(\X_i)I{\{\X_i\in R_{n,k}^S\}}.
$$
Therefore, $g(Y_i)$ used in the original formulation is
replaced with $\psi_n(\X_i)$ which serves to estimate
the conditional mean of $g(Y_i)$.  Given rules
$\zeta_{n,1},\ldots,\zeta_{n,K_n}$, the
modified VarPro measure of importance for $S$ is
$$
\Ds_n(S) = \sum_{k=1}^{K_n} W_{n,k}|\ts_{n}(\z_{n,k}^S)-
\ts_{n}(\z_{n,k})|
\label{E:varpro.modified}
$$
for prechosen weights $0\le W_{n,k}\le 1$,
$\sum_{k=1}^{K_n}W_{n,k}=1$. 

An important application of the modified procedure is survival
analysis. This is because trying to locally estimate the survival
function is difficult if right censoring is present. In such settings,
due to censoring, the observed data is $(T, \delta, \X)$, where $T =
\min(T^o, C^o)$ is the time of event, $\delta = I\{T^o \leq C^o\}$ is
the censoring indicator, and $T^o, C^o$ are the true survival and
censoring times. Because only $(\min(T^o, C^o), \delta, \X)$ is
observed, the time of event is only observed if censoring is not
present. The survival function is $S(t \mid \X) = \PP\{T^o > t \mid
\X\}$, and this corresponds to $\psi(\X) = \E(g(T^o) \mid \X)$ where
$g(T^o) = I\{T^o > t\}$. Let $\z$ be a rule and $R(\z)$ its
region. Because $T^o$ is potentially unobserved, we cannot estimate
$S(t \mid \X)$ locally using an estimator like
$$
\that_n(\z)
:=\frac{\sum_{i=1}^n  g(Y_i)1_{\{\X_i\in R(\z)\}}}{\sum_{i=1}^n I{\{\X_i\in R(\z)\}}}
= \frac{\sum_{i=1}^n  I{\{T_i^o>t, \X_i\in R(\z)\}}}{\sum_{i=1}^n I{\{\X_i\in R(\z)\}}}.
$$
If we want to use a sample average of the observed data, as in the
original VarPro formulation, then we have to account for
censoring.  For example, under the assumption of conditional
independence, 
one approach could be to use an IPCW (inverse of probability of censoring
weighting) estimator such as
$$
\that_n(\z)
= \(\sum_{i=1}^n I{\{\X_i\in R(\z)\}}\)^{-1}
\sum_{i=1}^n{\displaystyle
    \frac{\d_i}{G(T_i)}\Big[I{\{T_i>t, \X_i\in R(\z)\}}\Big]}
$$
where $G(u)=\PP\{C>u\}$ is the unknown censoring distribution.
However, there are known issues with IPCW estimators, such
as estimation for $G$, and problems with inverse weights becoming
large~\citep{Graf1999, Gerds2006}. 

As another example, consider variable selection for the restricted
mean survival time (RMST)~\citep{irwin1949standard,
andersen2004regression, royston2011use,kim2017restricted}.  The RMST is a
useful quantity summarizing lifetime behavior and is defined as the
survival function integrated from $[0,\tau]$
\Eq
\int_0^\tau S(t|\X)\,dt
=\E\[\int_0^\tau 1_{{\{T^o>t\}}}\,dt \Big|\X\]
=\E\[\int_0^{T^o \wedge \tau}dt \Big|\X\]
= \E\[T^o \wedge \tau|\X\].
\label{rmst}
\EndEq
The time horizon $\tau$ is usually selected to represent a
clinically meaningful endpoint, such as 1 year or 5 year survival.
Just like the survival
function due to censoring, the RMST is difficult to estimate locally.
Therefore, using an external estimator is a way to
overcome this issue. 

The following result shows that the modified VarPro procedure is
consistent under the assumption (A5) that $\psi_n$ converges uniformly to
$\psi$.  While this might seem to be a strong assumption, convergence
only has to hold over a suitably defined subspace (see~\cref{appF} for
further discussion regarding (A5)).

\begin{theorem}\label{varpro.external.noise.theorem}
The conclusions of
\cref{varpro.noise.theorem} and \cref{varpro.signal.theorem} hold
for $\Ds_n(S)$ under their stated conditions if in addition assumption
(A5) of~\cref{appF} holds.
\end{theorem}

In the next subsections we present examples illustrating the modified
procedure applied to right censored survival settings.

\subsection{High-dimensional, low sample size, variable selection for survival}

For our first illustration we consider a high-dimensional survival
simulation.  The survival time $T^o$ was simulated by
$$
T^o= \log \[1 + V \exp\(\sum_{j=1}^{p} \beta_j \left\{\Xs{j}\right\}^2\)\],\hskip5pt
V\sim \frac{4}{10}\exp(.5)+\frac{1}{10}\exp(1)
+\frac{2}{10}\exp(1.5)+\frac{3}{10}\exp(3)
$$
where $V$ was sampled independently of $\X$ and is a four-component
mixture of exponential random variables with rate parameters (inverse
means) $.5, 1, 1.5, 3$.  The first $p_0=10$ features of $\X$ were
signal variables: these were assigned coefficient values
$\beta_j=\frac{1}{2}\log(1+\sqrt{p/p_0})$.  All features had marginal
uniform $U(0,1)$ distributions.  The noise features were uncorrelated;
the signal features had pairwise correlation $\rho=3/7$.  Random
censoring at 50\% and 75\% rates were used.  A sample size $N=200$ was
used while varying $p$.

For $\psi$, we used the integrated cumulative hazard
function (CHF) and for the external estimator $\psi_n$, we chose
random survival forests (RSF)~\citep{Ishwaran2008} using 
the \rfsrc\, package~\citep{rfsrc}.  
Because the ensemble CHF is piecewise constant, no approximation was
needed to integrate it.

Comparison procedures included Cox regression with lasso
penalization (coxnet) using the \glmnet\,
package~\citep{simon2011regularization} and
BC-VIMP and minimal depth
(MD)~\citep{ishwaran2010high, ishwaran2011random}
using the \rfsrc~\citep{rfsrc} R-package.
These methods were specifically selected due to their proven
effectiveness in feature selection for high-dimensional survival
problems. Given the complex nature of the simulations, careful
tuning was necessary for all methods. For coxnet, the default
one-standard error rule from glmnet for choosing the lasso
regularization parameter $\l$ during 10-fold validation resulted in
overly sparse solutions. Therefore, we opted for the $\l$ that
produced the smallest out-of-sample error (the minimum rule).
For minimal
depth, we used the mean minimal depth threshold
value~\citep{ishwaran2010high, ishwaran2011random}.
For VarPro, the $Z_0$ cutoff value was obtained using an
out-of-sample technique similar to the regression and classification
benchmark experiments of Section 4.  Specifically, sequential models used to
select the optimized cutoff value were fit using RSF.  Out-of-sample
performance was evaluated using the continuous rank probability score
(CRPS)~\citep{Graf1999, Gerds2008}.

Simulations were repeated 100 times independently.  Feature selection
was assessed by gmean, precision and accuracy.  The results were
averaged across the runs and are given in~\cref{survTab}.
Similar to our previous benchmark experiments for regression and
classification, we observe that VarPro achieves the best overall
performance in all metrics (adjusted $p$-value < 0.00001)

\begin{table}[phtb]
\caption{\small \label{survTab}High-dimensional survival simulation ($N=200$,
  $p=500,1000,1500$ and $p_0=10$) where signal variables are
  correlated.  Random censoring was applied at 50\% and 75\% rates. Results
  are averaged over 100 independent runs.}
\vspace{10pt}
\centering
\fbox{%
\begin{tabular}{cc|rrr|rrr}
\multicolumn{2}{c}{}&
\multicolumn{3}{c}{50\% Censoring}&
\multicolumn{3}{c}{75\% Censoring}\\
&$p$ & Gmean & Precision & Accuracy 
   & Gmean & Precision & Accuracy \\
  \hline\\[-10pt]
 & 500 & 0.91 & 0.49 & 0.97 & 0.77 & 0.42 & 0.97 \\ 
coxnet & 1000 & 0.93 & 0.42 & 0.98 & 0.79 & 0.36 & 0.98 \\ 
   & 1500 & 0.94 & 0.39 & 0.99 & 0.79 & 0.32 & 0.99 \\ 
   \hline
 & 500 & 0.95 & 0.55 & 0.98 & 0.87 & 0.42 & 0.97 \\ 
VarPro & 1000 & 0.96 & 0.53 & 0.99 & 0.91 & 0.43 & 0.99 \\ 
   & 1500 & 0.97 & 0.57 & 0.99 & 0.92 & 0.43 & 0.99 \\ 
   \hline
 & 500 & 0.86 & 0.07 & 0.74 & 0.81 & 0.06 & 0.70 \\ 
BC-VIMP & 1000 & 0.89 & 0.05 & 0.79 & 0.84 & 0.04 & 0.75 \\ 
 & 1500 & 0.91 & 0.04 & 0.82 & 0.86 & 0.03 & 0.79 \\ 
  \hline
 & 500 & 0.88 & 0.37 & 0.80 & 0.77 & 0.41 & 0.81 \\ 
MD & 1000 & 0.88 & 0.43 & 0.82 & 0.72 & 0.38 & 0.79 \\ 
  & 1500 & 0.89 & 0.52 & 0.84 & 0.67 & 0.52 & 0.83  
\end{tabular}}
\end{table}

\subsection{Heart rate recovery long term survival}

Exercise stress testing is commonly used to assess patients with known
or suspected coronary artery disease.  A useful predictor of mortality
is fall in heart rate immediately after exercise stress testing, or
heart rate recovery~\citep{imai1994vagally}.  Heart rate recovery is
defined as the heart rate at peak exercise minus the heart rate
measured 1 minute later.  The hypothesis that heart rate recovery
predicts mortality has been tested and validated in a number of
cohorts~\citep{cole2000heart}.  Here we study this issue by
 considering how predictive heart rate recovery is in the presence of
other potentially useful features by making use of the modified VarPro
procedure.

For this analysis we use data from the study 
in~\cite{ishwaran2004relative}.  The study
considered $N=$23,701 patients
referred for symptom-limited exercise testing. Each patient underwent
an upright cool-down period for the first 2 minutes after recovery.
Detailed data regarding reason for testing, symptoms, cardiac risk
factors, other medical diagnoses, prior cardiac and noncardiac
procedures, medications, resting electrocardiogram, resting heart
rate, and blood pressure were recorded prospectively prior to
testing. During each stage of exercise, and during the first 5 minutes
of recovery, data were recorded regarding heart rate, blood pressure,
ST-segment changes, symptoms, and arrhythmias.  In total $p=85$ variables
were available for the analysis.  All cause mortality was used for the
survival outcome.  Data was right-censored with mean follow-up 
of 5.7 years (range .75 to 10.1 years) during which 1,617
patients died.

\begin{figure}[phtb]
\vskip10pt
  \begin{center}
\resizebox{5.20in}{!}{\includegraphics{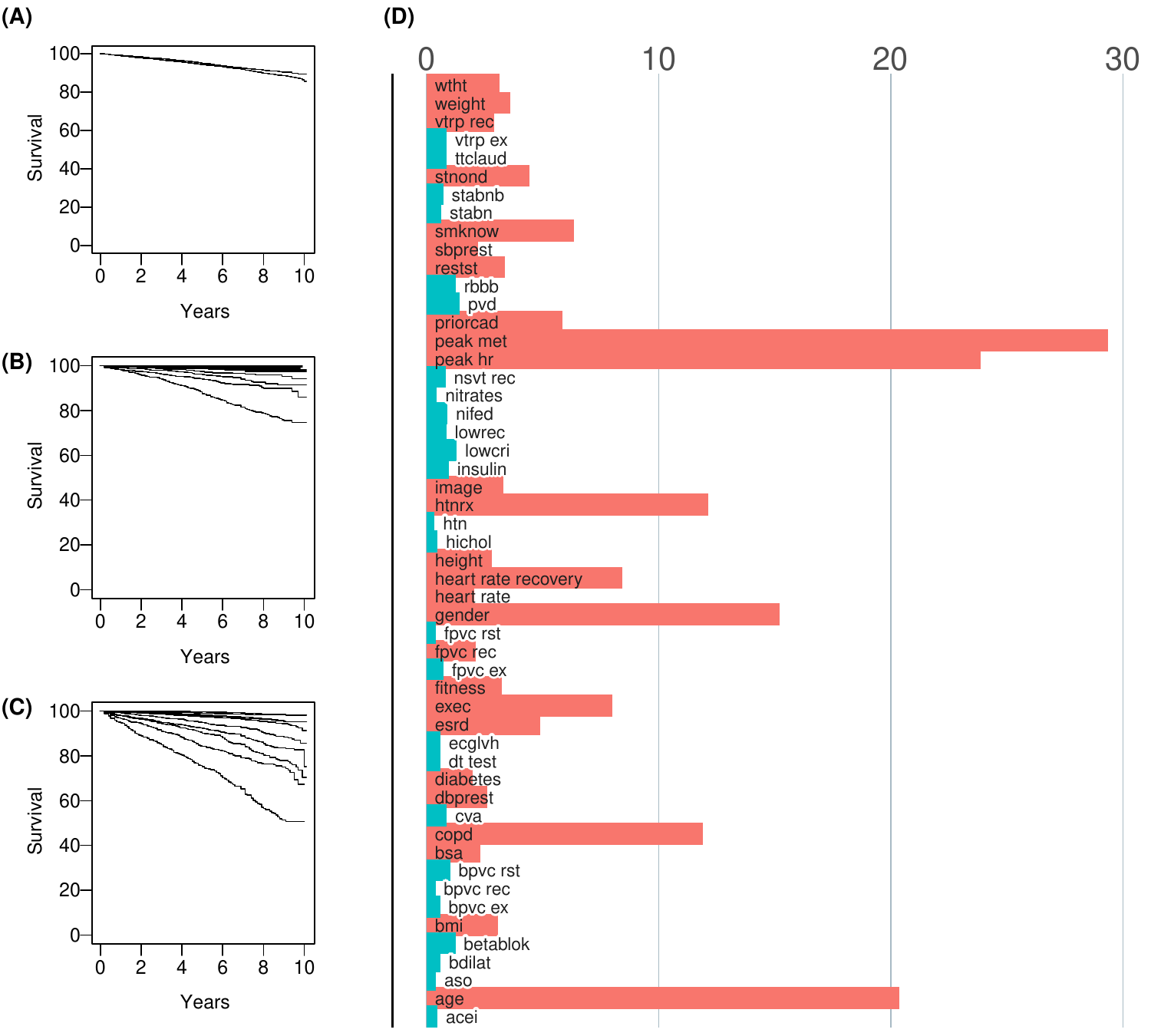}}
\end{center}
\caption{\small Results of heart rate recovery survival analysis using
  the modified VarPro procedure.  (A) Survival for men and
  women; (B) Survival for women, stratified by peak met; (C) Survival
  for men, stratified by peak met; (D) Standardized importance for
  variables (red are signal, blue are noise variables).}
\label{hrr}
\end{figure}

For $\psi$, we use the RMST evaluated at $\tau=3$ years.  The RMST was
calculated by~\mref{rmst} using the estimated survival function from a
RSF analysis.  As before, calculations used the
\rfsrc\, package~\citep{rfsrc}.  The standardized importance values
from VarPro are given in~\cref{hrr}~(D).  Signal variables (red) and
noise variables (blue) are identified using a $Z_0$ cutoff value where
the latter was calculated using an out-of-sample strategy as in the
previous section.  From the figure, we can observe that heart
rate recovery is identified as a signal variable.  However, at the
same time, we also observe several other significant variables,
some with even larger importance values, such as age, peak
met, peak heart rate and sex.

For comparison, we re-analyzed the data using tree boosting with Cox
regression.  Computations were run using the R-package
\gbm~\citep{gbmr}.  The top 5 variables from boosting were age, peak
met, heart rate recovery, COPD and heart rate, which generally agree
with the top variables found by VarPro.  However, an interesting
difference between the two procedures was their ranking for sex.
VarPro ranked sex as among its top 5 variables whereas sex did not
even make it into the top 12 variables for boosting.  This is
important because sex is a variable that is often overlooked in the
heart failure literature.

Further analysis is presented with separate Kaplan-Meier survival
curves for men and women in~\cref{hrr} (A), showing minimal
differences. However, when stratifying peak met (a significant
variable) into deciles and plotting survival curves for each category,
a notable distinction emerges. For men (C), survival decreases more
significantly than for women (B) at lower peak met values (survival
curves descending from top to bottom). This suggests that men are more
adversely affected by reduced peak met, implying greater sensitivity
to exercise capacity, as peak met is an indicator of this.  These
observations are consistent with prior research
by~\cite{hsich2007long}, which indicated that women have a lower
mortality risk than men at any given level of exercise or oxygen
capacity.

\section{Discussion}

We demonstrated that VarPro performs well in many settings. In a large
cardiovascular survival study (Section 5), VarPro was able to identify
a variable without a main effect but with a strong interaction, which
led to discovering a meaningful difference in long-term survival for
patients at risk for heart disease. This is impressive because
identifying variables with no main effects but significant
interactions is generally considered difficult, even for the very best
variable selection methods. The ability to deal with interactions was
also evident in the synthetic regression experiments (Section 4),
where models frequently included interaction terms. VarPro not only
performed well but also stood out in scenarios with correlated
features, which is valuable since real-world data often involves
complex interactions and correlations. Section 4 provided further
illustration of the ability to handle real-world correlation using a
large benchmark analysis of microarray datasets.

\begin{figure}[phtb]
  \centering
  \vskip10pt
\resizebox{6in}{!}{\includegraphics[page=1]{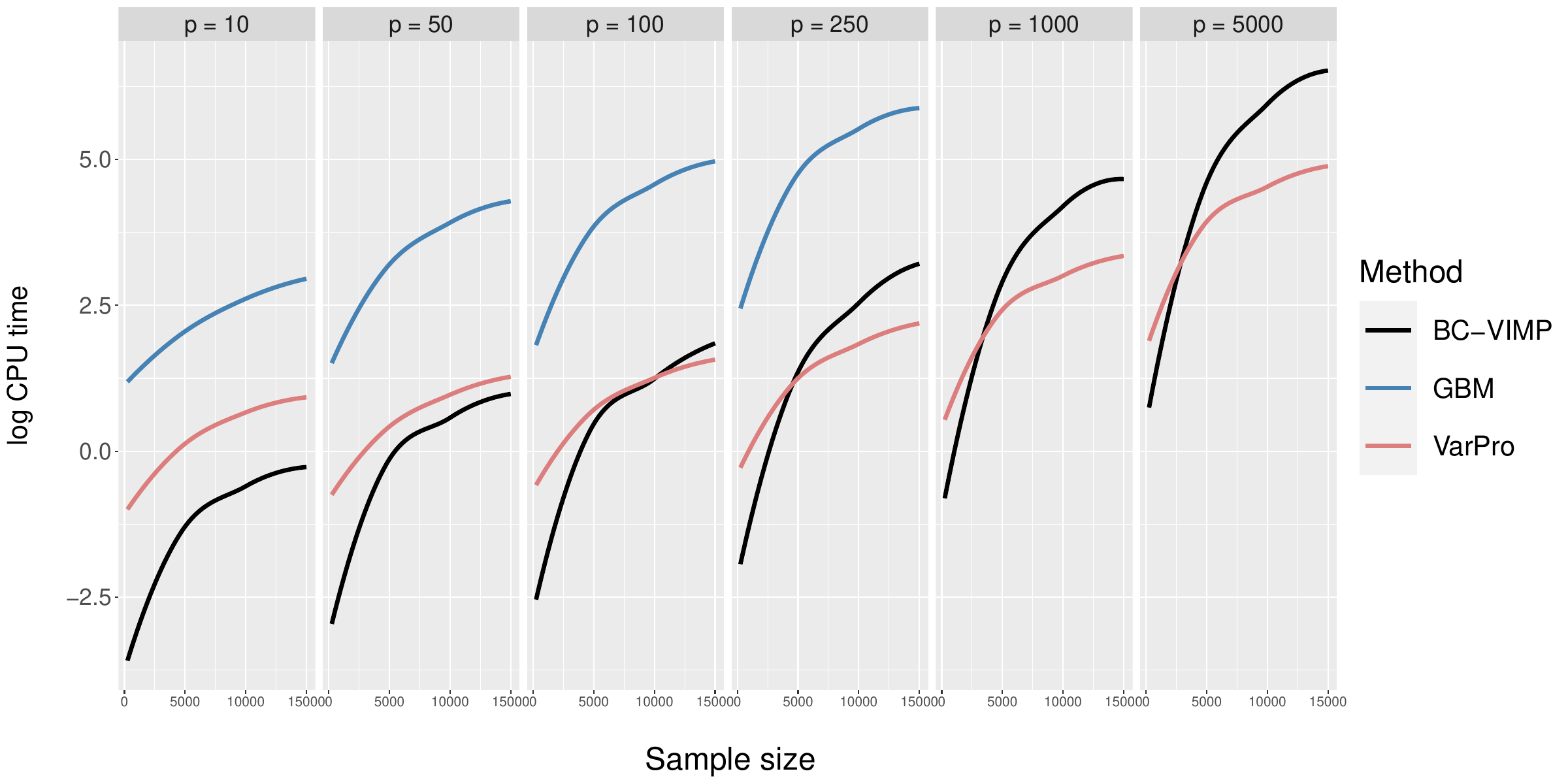}}
  \caption{\small Log CPU times in seconds for VarPro (red lines)
    using the Friedman~1
    simulation (\cref{appE}) with varying sample size $N$ and dimension $p$.  Shown
    for comparison are log CPU times for BC-VIMP (black
    lines) using the \ranger\, R-package (500 trees 
    used for each forest) and gradient boosting (blue lines) using the
    \gbm\, R-package (the optimal number of trees used in each boosted
    experiment was determined by 10-fold
    validation with a maximum value of 2000 trees).}
\label{cputime}
\end{figure}

Another strength of VarPro is that it makes use of rule-based
selection. This replaces the problem of building a high-performance
model~\citep{modelfree,additivevs} with a series of lower-dimensional
localized variable selection problems that can be solved
computationally fast. All the examples presented in this paper can be
computed efficiently, and in our experience, we have found VarPro to
be very fast. As an illustration, \cref{cputime} shows the CPU times
for VarPro (red lines) across different simulation sizes from $N=250$
to $N=15,000$ and dimensions from $p=10$ to $p=5000$ using the
Friedman~1 regression simulation (\cref{appE}). For context, CPU times
for BC-VIMP (black lines) using the R-package
\texttt{ranger}~\citep{ranger} and for gradient boosting (blue lines)
using the R-package \texttt{gbm}~\citep{gbmr} are also displayed,
noting that gradient boosting times were limited by sample sizes and
dimensions due to long run time durations. VarPro's run times prove to
be more favorable as $N$ and $p$ increase, highlighting its
computational efficiency and viability in handling large datasets
compared to existing methods.

Despite the encouraging results, the method can still be enhanced. A
potential area for improvement is rule generation. Our approach
utilized trees guided by split-weights derived from combining lasso
with shallow trees. Although effective in this context, the
development of automated strategies for generating high-performing
rules needs additional exploration. Nonetheless, a strength of our
methodology lies in providing a theoretical road map for what needs to
be done. The rule generation step needs to guarantee that regions for
signal variables shrink to zero, but as has been pointed out, such an
assumption seems reasonable. Given the minimal assumptions on rule
construction, this approach should facilitate further adaptations and
applications.

Another potential limitation is identification of signal variables in
scenarios where a unique Markov boundary may not exist.
As discussed by~\cite{statnikov2013algorithms}, there can be multiple
Markov boundaries of the response variable.  In other words, $\psi$
may not be identified because several variables could interchangeably
provide the same amount of information for $\psi$.  Since our current
discovery algorithm is to quantify the relative importance of
predictors one by one, we tend to select the signal variables as the
union set of multiple Markov boundaries that contain redundant
information. However, in this case, our method retains its property
of being able to filter noise variables.

We use the following simulation for
illustration.   Data was drawn from the regression model
$$
Y = \Xs{1} + \Xs{2} +\e,
\hskip10pt \e\sim\text{N}(0,1).
$$
The feature vector is of dimension $p=150$ and drawn from a multivariate
normal with zero mean and equicorrelation matrix with correlation
parameter $\rho=.4$.  Variables $\Xs{3}$ and $\Xs{4}$ were
modified so that 
$$
\Xs{3}=\beta\Xs{1}+(1-\beta)\Xs{2},\hskip5pt
\Xs{4}=(1-\beta)\Xs{1}+\beta\Xs{2}.
$$
Therefore, this reflects a regression setting where some columns are linearly
related to others.  Notice that
$$
\psi(\X)
= \E(Y|\X)
= \Xs{1} + \Xs{2}
= \Xs{3} + \Xs{4}.
$$
Therefore according
to~\cref{signal.noise.definition}, the signal variables must
be $\ss=\{1,2,3,4\}$.  It cannot be $\ss=\{1,2\}$, since we
would have $\psi(\X)=\psi(\X^{\setminus(\ss)})=\psi(\X^{\{3,4\}})$.
Likewise it cannot be $\ss=\{3,4\}$,
since then
$\psi(\X)=\psi(\X^{\setminus(\ss)})=\psi(\X^{\{1,2\}})$.
Therefore $\ss$ must contain all 4 variables, creating a redundant
scenario.

For demonstration, VarPro was executed 1000 times independently using
the above simulation, with a sample size of $N$=1500 and $\beta = 1/4$. The
first row of~\cref{markov}  lists the averaged standardized importance for
the first 4 variables, while all remaining variables are grouped under
the “noise” category. The second row shows the frequency of each
variable's selection. The results highlight VarPro's capabilities:
(1) efficiently weeding out noise variables and (2) identifying
redundant signal variables, confirming the earlier statements about
VarPro's performance in scenarios with non-unique Markov boundaries.

\begin{table}[phtb]
  \caption{\small\label{markov} Regression simulation with non-unique Markov boundary.
  The first row lists VarPro standardized importance values.
  The second row lists the selection frequency for each variable.
  The conditional mean can be described by both $\{\Xs{1},\Xs{2}\}$
  and $\{\Xs{3},\Xs{4}\}$, therefore there is redundancy in the
  definition of a signal variable.
  Variables $\Xs{j}$ for
  $j=5,\ldots,150$ are combined into category ``noise''.}
  \vspace{10pt}
  \centering
\begin{tabular}{rrrrrr}
  \hline
 & $\Xs{1}$ & $\Xs{2}$ & $\Xs{3}$ & $\Xs{4}$ & noise \\ 
  \hline
Importance & 5.63 & 5.61 & 10.30 & 10.29 & 0.00 \\ 
  Percent & 98.70 & 99.10 & 100.00 & 100.00 & 0.03 \\ 
   \hline
\end{tabular}
\end{table}


\begin{appendix}

\section{Uniform approximation for the average size of a region}\label{appA}

The following lemma will be used in several places and provides a
uniform approximation for the average sample size of a region.
The notation $b_n\Gt a_n$ is used to signify
$b_n/a_n\rightarrow\infty$.

\begin{lemma}\label{hoeffding.lemma}
Let $R_{n,k}\subseteq\xx$ be a collection of sets such that $\PP\{R_{n,k}\}>0$,
$k=1,\ldots,K_n$, and define
$m_{n,k}=\sum_{i=1}^n I\{\X_i\in R_{n,k}\}$ where $\X_1,\ldots,\X_n$
are i.i.d.\ random vectors over $\xx$.  If  $K_n\le O(\log n)$
and $m_{n,k}\ge m_n\Gt
\sqrt{n\log\log n}$, then the following identity holds
over a set with probability
tending to one uniformly over $k=1,\ldots,K_n$:
\Eq
\frac{n}{m_{n,k}} =
\frac{1}{\PP(R_{n,k})}\Big(1+\xi_{n,k}^*\Big),\hskip8pt\text{where }\,
|\xi_{n,k}^*|\le \dfrac{\displaystyle{\sqrt{\log\log n}}}{\displaystyle{n^{-1/2}m_n}}
\rightarrow 0.
\label{hoeffding.mnk}
\EndEq
\end{lemma}

\vskip10pt\begin{proof}
Observe that $m_{n,k}=\sum_{i=1}^n I\{\X_i\in R_{n,k}\}$ is a sum of
i.i.d.\ Bernoulli random variables.  Therefore by Hoeffding's
inequality~\citep{hoeffding1963probability},
$$
\PP\left\{\max_{1\le k\le K_n}\left|n^{-1}m_{n,k} - \PP(R_{n,k})\right|\ge \e\right\}
\le 2K_n\exp\(-2\e^2 n\).
$$
Let $\aa_n^c$ be the event inside the probability.  If 
$\e$ is allowed to depend on $n$ so that $2\e^2n\ge 2(\log \log n)$, then
the bound on the right is order $K_n (\log n)^{-2}\rightarrow 0$.  Therefore
over $\aa_n$, an event which occurs with probability tending to one, 
$$
|\xi_{n,k}|\le \xi_n,\hskip10pt\text{where }
\xi_{n,k}= \PP(R_{n,k})-n^{-1}m_{n,k},\hskip3pt
\xi_n=n^{-1/2}\sqrt{\log\log n}.
$$
Using $m_{n,k}\ge m_n$, then with probability tending to
one over $\aa_n$:
$$
\frac{n}{m_{n,k}} =
\frac{1}{\PP(R_{n,k})}
+ \frac{\xis{n,k}}{\PP(R_{n,k})},\hskip10pt\text{where }
\Bigg|\xis{n,k}:=\frac{\xi_{n,k}}{n^{-1} m_{n,k}}\Bigg|
\le\frac{\xi_n}{n^{-1}m_n}.
$$
Upon rearrangement this gives~\mref{hoeffding.mnk}.
\end{proof}

\section{Uniform approximation for sample averages with varying size}\label{appB}

In order for $\D_n(S)$ to have good theoretical performance,
$\that_n(\z_{n,1}),\ldots,\that_n(\z_{n,K_n})$ and the released values
$\that_n(\z_{n,1}^S),\ldots,\that_n(\z_{n,K_n}^S)$ must be controlled uniformly.
This imposes a limit on the sequences $K_n, m_n(\z_{n,k})$ and
$m_n(\z_{n,k}^S)$.   We describe a lemma for
this purpose.

First we rewrite the VarPro estimators in a
slightly different way.  
For a given rule $\z$, let $b(\z)=\E[\psi(\X)I\{\X\in R(\z)\}]$ where
recall that $\psi(\X)=\E(g(Y)|\X)$.  Notice that $\that_n$ can be
rewritten as
\Eq
\that_n(\z)
=
\frac{1}{m_n(\z)}\sum_{i=1}^n Z_i(\z)
+ \frac{nb(\z)}{m_n(\z)},
\label{expansion.prelim}
\EndEq
where $Z_i(\z)=g(Y_i)I\{\X_i\in R(\z)\} - b(\z)$ are
i.i.d.\ random variables with a mean of zero:
\Array
\E(Z_i(\z))
&=& \E\biggl\{\E\Bigl[(g(Y) - \psi(\X)I\{\X\in
  R(\z)\}\Big|\X\Bigr]\biggr\}\\
&=& \E\biggl\{I\{\X\in R(\z)\}\E\Bigl[(g(Y) -
  \psi(\X))\big|\X\Bigr]\biggr\}
=0.
\EndArray
Recall that $m_{n,k}=m_n(\z_{n,k})$,
$m_{n,k}^S=m_n(\z_{n,k}^S)$ and $R_{n,k}=R(\z_{n,k})$,
$R_{n,k}^S=R(\z_{n,k}^S)$.  Applying a similar centering as
in~\mref{expansion.prelim} to the released rule, we have
\EqArray
\that_n(\z_{n,k}^S)
-\that_n(\z_{n,k})
&=&
\[\frac{1}{m_{n,k}^S}\sum_{i=1}^n Z_i(\z_{n,k}^S) 
- \frac{1}{m_{n,k}}\sum_{i=1}^n Z_i(\z_{n,k})\]
\nonumber\\
&&+
\[\frac{n b(\z_{n,k}^S)}{m_{n,k}^S}-\frac{n b(\z_{n,k})}{m_{n,k}}\],
\label{expansion}
\EndEqArray
where  (similar definitions apply to $\z_{n,k}^S$):
$$
Z_i(\z_{n,k})  = g(Y_i)I\{\X_i\in R_{n,k}\}-b(\z_{n,k}),\hskip10pt
b(\z_{n,k})    = \E\[\psi(\X)I\{\X\in R_{n,k}\}\].
$$
The sums appearing in~\mref{expansion} will be
shown to converge to zero uniformly using the lemma given next.
The quantity in the second square bracket will be dealt with later.
It represents a ``bias'' term that is asymptotically zero for noise
variables but non-zero for signal variables.  

\subsection{Key lemma for uniform approximation of sample averages}

For each $n$, let $\Zn_{1,k},\ldots,\Zn_{n,k}$ be independent random
variables such that $\E(\Zn_{i,k})=0$ and $\E[(\Zn_{i,k})^2]\le\s^2<\infty$
for $i=1,\ldots,n$ and $k=1,\ldots,K_n$.  Let 
$$
S_{n,k}=\sum_{i=1}^n \Zn_{i,k},
\hskip5pt\text{and}\hskip5pt
T_{n,k}
=\frac{1}{M_{n,k}} \sum_{i=1}^n \Zn_{i,k}
=\frac{1}{M_{n,k}} S_{n,k}
$$
where $M_{n,k}$ are random values (not necessarily independent of
$\Zn_{1,k},\ldots,\Zn_{n,k}$) satisfying $n\ge M_{n,k}\ge M_n>0$ for
$k=1,\ldots,K_n$.  We wish to identify conditions for the deterministic
sequences $M_n,K_n$ such that 
$T_{n,k}$ converges to zero uniformly over $k=1,\ldots,K_n$
as $n\rightarrow\infty$.

Let $L_n=(n K_n)^{-1/2}$.  Then
by Chebyshev's inequality, for any constant $C>0$,
\Eq
\PP\left\{L_n\bigl|S_{n,k}\bigr|\ge C \right\}
=\PP\left\{\biggl|\sum_{i=1}^n \Zn_{i,k}\biggr|\ge \frac{C}{L_n} \right\}\\
\le\frac{L_n^2}{C^2}\sum_{i=1}^{n} \E\[(\Zn_{i,k})^2\]\le \frac{\s^2}{C^2 K_n}.
\label{chebyshev}
\EndEq
Let $\bb_{n,k}=\{\omega: |L_n S_{n,k}|\ge C\}$ and
$\bb_n=\bigcup_{k=1}^{K_n} \bb_{n,k}$.  Because
$M_{n,k}\ge M_n$, setting
$\d_n=L_n^{-1}M_n^{-1}=M_n^{-1}(nK_n)^{1/2}$ we obtain
\Eq
|T_{n,k}|
=
\frac{L_n^{-1}}{M_{n,k}}|L_nS_{n,k}|\, I_{\bb_n^c}
+ |T_{n,k}|\, I_{\bb_n}\\
\le C\d_n + \max_{1\le k\le K_n}|T_{n,k}| \, I_{\bb_n}.
\label{Dkn.bound}
\EndEq
For the first term on the right of~\mref{Dkn.bound} to converge,
$M_n$ must converge at a rate faster than $(nK_n)^{1/2}$.
For the second term, using~\mref{chebyshev}, observe that
\Array
\PP(\bb_n)
&=&
\PP\(\bigcup_{k=1}^{K_n} \bb_{n,k}\)\\
&=&
\PP\left\{\max_{1\le k\le K_n}\,|L_n S_{n,k}|\ge C \right\}\\
&\le&
\sum_{k=1}^{K_n}\PP\left\{|L_n S_{n,k}|\ge C \right\}
\le \frac{K_n\s^2}{C^2K_n}=\frac{\s^2}{C^2}.
\EndArray
Therefore using Markov's inequality, we have for each $\e>0$,
\Array
\PP\left\{\max_{1\le k\le K_n}|T_{n,k}|\, I_{\bb_n} \ge \e\right\}
&\le&
\frac{\sum_{k=1}^{K_n}\E\(|T_{n,k}|I_{\bb_n}\)}{\e}\\
&\le&
\dfrac{{\displaystyle\sum_{k=1}^{K_n}\sqrt{\E\(T_{n,k}^2\)\PP(\bb_n)}}}{\e}
\hskip20pt\text{(Cauchy-Schwarz)}\\[5pt]
\\
&\le&\frac{\s}{C\e}\sum_{k=1}^{K_n}\sqrt{\E\(T_{n,k}^2\)}
\hskip34pt\text{(bound from $\PP(\bb_n)$ above)}.
\EndArray
Because $\E(T_{n,k}^2)\le M_{n}^{-2}\sum_{i=1}^n\E\[(\Zn_{i,k})^2\]\le
M_n^{-2}n\s^2$,
$$
\PP\left\{\max_{1\le k\le K_n}|T_{n,k}|\, I_{\bb_n} \ge \e\right\}
\le \frac{\s^2n^{1/2}K_n}{C\e M_n}.
$$
Therefore if $M_n^{-1}n^{1/2}K_n\rightarrow 0$, then
by~\mref{Dkn.bound} we have shown
$$
|T_{n,k}|\le C\d_n+o_p(C^{-1}\s^2 M_n^{-1}n^{1/2}K_n)
$$
uniformly over $k$.  But notice that $M_n^{-1}n^{1/2}K_n=
K_n^{1/2}\d_n\ge C\d_n$ eventually, thus we
have proven the following lemma.

\begin{lemma}\label{key.lemma}
If $M_n\uparrow\infty$ such that
$M_n^{-1}n^{1/2}K_n\rightarrow 0$ then $|T_{n,k}|\le o_p(1)$
uniformly over $k=1,\ldots,K_n$.
\end{lemma}

\section{Consistency for noise variables (proof of Theorem~\ref{varpro.noise.theorem})}\label{appC}

The idea for the proof is as follows.
The two sums in the first square bracket of~\mref{expansion} will be dealt
with by using~\cref{key.lemma} (set $M_{n,k}=m_{n,k}$,
$M_n=m_n$ and $\Zn_{i,k}=Z_i(\z_{n,k})$, then $T_{n,k}$ in
\cref{key.lemma} equals $m_{n,k}^{-1}\sum_{i=1}^n Z_i(\z_{n,k})$).
The term inside the second square bracket of~\mref{expansion} is a
bias term and will be shown to be
asymptotically equal to
\Eq
\b_{n,k}(S)=\E(\psi(\Xbs{\ss})|\,\X\in R_{n,k}^S) -\E(\psi(\Xbs{\ss})|\,\X\in R_{n,k}).
\label{bias.asymptotic}
\EndEq
This will be shown to converge to zero for noise
variables by using the smoothness assumptions for $\psi$ and other
conditions assumed by the theorem.

\vskip10pt\begin{proof}
Apply~\cref{key.lemma} to each of the sums in the first square
bracket of~\mref{expansion}.  The lemma applies since
$\{Z_i(\z_{n,k}),Z_i(\z_{n,k}^S)\}$ are centered i.i.d.\ variables
with bounded second moment.  The latter holds by Assumption (A1).  For
$\{Z_i(\z_{n,k})\}$, let $M_{n,k}=m_{n,k}, M_n=m_n$ where notice
that $m_n^{-1}n^{1/2}K_n\rightarrow 0$ when $K_n\le
O(\log n)$ and $m_n=n^{1/2}\g_n$ where $\g_n\Gt\log n$; thus verifying
the rate condition of the lemma.  Moreover, because
$m_{k,n}^S\ge m_{k,n}$ since
$R_{n,k}\subseteq R_{n,k}^S$, the conditions of the lemma also
hold for $\{Z_i(\z_{n,k}^S)\}$ with $M_{n,k}=m_{n,k}^S, M_n=m_n$.

Therefore by~\cref{key.lemma}, which holds uniformly,
\EqArray
\D_n(S)
&\le&
\sum_{k=1}^{K_n}W_{n,k}|o_p(1) + b_{n,k}|\nonumber\\
&\le& o_p(1) +
\sum_{k=1}^{K_n} W_{n,k}|b_{n,k}|,
\hskip10pt\text{uniformly},
\label{delta.bound.null.1}
\EndEqArray
where (see the second term of~\mref{expansion} and use $\psi(\X)=\psi(\Xbs{\ss})$)
\Eq
b_{n,k}
=\frac{n}{m_{n,k}^S} \E[\psi(\Xbs{\ss})I\{\X\in R_{n,k}^S\}]
- \frac{n}{m_{n,k}} \E[\psi(\Xbs{\ss})I\{\X\in R_{n,k}\}].
\label{bias.noise.bound.1}
\EndEq
Using~\cref{hoeffding.lemma} we will show~\mref{bias.noise.bound.1} 
approximates
\EqArray
&&\hskip-45pt
\frac{\E[\psi(\Xbs{\ss})I\{\X\in R_{n,k}^S\}]}{\PP(R_{n,k}^S)} 
- \frac{\E[\psi(\Xbs{\ss})I\{\X\in R_{n,k}\}]}{\PP(R_{n,k})}\nonumber\\
&&=
\E(\psi(\Xbs{\ss})|\,\X\in R_{n,k}^S)-
\E(\psi(\Xbs{\ss})|\,\X\in R_{n,k})\nonumber\\[3pt]
&&:=
\E_{n,k}^S(\psi)-
\E_{n,k}(\psi),
\label{bias.noise.bound.2}
\EndEqArray
where for notational simplicity we write $\E^S_{n,k}$ and $\E_{n,k}$
for the conditional expectation of $\X$ given $R_{n,k}^S$ and $R_{n,k}$.  Observe
that~\mref{bias.noise.bound.2} is the asymptotic bias $\b_{n,k}:=\b_{n,k}(S)$
discussed earlier in~\mref{bias.asymptotic}.

Apply~\cref{hoeffding.lemma} noting $m_{n,k}\ge m_n=n^{1/2}\g_n
\Gt \sqrt{n\log\log n}$.  By~\mref{hoeffding.mnk}, there
exists a set $\aa_n$ with probability tending to one, such that
$$
\frac{n}{m_{n,k}} =
\frac{1}{\PP(R_{n,k})}
+ \frac{\xis{n,k}}{\PP(R_{n,k})},\hskip10pt\text{where }
|\xi_{n,k}^*|\le
\xi_n^*= \g_n^{-1}\sqrt{\log\log n}\rightarrow 0.
$$
In a similar fashion, using $m_{n,k}^S\ge m_{n,k}\ge m_n=n^{1/2}\g_n$,
there exists a set $\aa_{n}^S$ with probability tending to
one, such that
$$
\frac{n}{m_{n,k}^S} =
\frac{1}{\PP(R_{n,k}^S)}
+ \frac{\xiS{n,k}{S}}{\PP(R_{n,k}^S)},\hskip10pt\text{where }
|\xiS{n,k}{S}|\le \xi_n^*.
$$
Thus from~\mref{bias.noise.bound.1}, 
over the set $\aa_n\bigcap\aa_n^S$ (an event with probability
tending to 1), we have
\Eq
|b_{n,k}|
=
\Big|\b_{n,k}
+ \xiS{n,k}{S}\E_{n,k}^S(\psi)
- \xis{n,k}   \E_{n,k}(\psi)\Big|
\le
|\b_{n,k}|
+ \xi_n^* \Big(|\E_{n,k}^S(\psi)| + |\E_{n,k}(\psi)|\Bigr).
\label{bias.noise.bound.3}
\EndEq

The smoothness assumption (A2) for $\psi$ and the shrinking condition
(A3) for $R_{n,k}$ are now used to expand $\E_{n,k}^S(\psi)$ and
$\E_{n,k}(\psi)$ to first order which will show~\mref{bias.noise.bound.2}
is asymptotically zero and will enable us to further
bound~\mref{bias.noise.bound.3}.  Let $\x_{n,k}$ be an arbitrary
point in $R_{n,k}$.  By the mean-value theorem,
for each $\x\in R_{n,k}$ there exists
a $\l_{n,k}\in[0,1]$, such that
$$
\psi(\x)-\psi(\x_{n,k})
=(\x-\x_{n,k})'\f(\x_{n,k}^*)
$$
where $\x_{n,k}^*=\x_{n,k} + \l_{n,k}(\x-\x_{n,k})$ (note
that the dependence of $\l_{n,k}$ on $\x$ is
suppressed). Using
$\f(\x_{n,k}^*)=\f(\x_{n,k})+[\f(\x_{n,k}^*)-\f(\x_{n,k})]$, the
Lipshitz condition~\mref{lipschitz.assume}, and keeping in mind $\f$
is zero over the coordinates for $\nn$,
$$
|\psi(\x)-\psi(\x_{n,k})|
\le |(\xss-\xss_{n,k})'\fs(\x_{n,k})|
+ C_0 |(\xss-\xss_{n,k})|'|(\x_{n,k}^{*(\ss)}-\xss_{n,k})|.
$$
Applying the Cauchy-Schwarz inequality to the first term on the right,
and using assumption (A3), we have for $\x\in R_{n,k}$
\EqArray
|\psi(\x)-\psi(\x_{n,k})|
&\le&
||(\xss-\xss_{n,k})||_2 ||\fs(\x_{n,k})||_2
+ C_0 |(\xss-\xss_{n,k})|'|(\x_{n,k}^{*(\ss)}-\xss_{n,k})|\nonumber\\
&\le&
\diam_{\ss}(R_{n,k})\Bigl[||\fs(\x_{n,k})||_2+ C_0\Bigr]\nonumber\\
&\le&
r_{n}\Bigl[||\fs(\x_{n,k})||_2+
  C_0\Bigr]=r_{n}\Bigl[||\f(\x_{n,k})||_2+ C_0\Bigr].
\label{psi.noise.bound}
\EndEqArray
Therefore
$\E_{n,k}(\psi) = \psi(\x_{n,k}) + r_{n,k}$, where
$$
\Big|r_{n,k}:=\E_{n,k}(\psi(\X)-\psi(\x_{n,k}))\Big|
\le r_n \Bigl[||\f(\x_{n,k})||_2+ C_0\Bigr]:=r_{n,k}^*.
$$
By a similar argument, $\E_{n,k}^S(\psi) = \psi(\x_{n,k}) +
r_{n,k}^S$ where $|r_{n,k}^S|\le r_{n,k}^*$ satisfies the same bound
as $r_{n,k}$.  This is because~\mref{psi.noise.bound} holds for $\x\in
R_{n,k}^S$ because
$R_{n,k}^S$ only differs from $R_{n,k}$ along the noise coordinates (since
$S\subseteq\nn$).

Therefore $\E_{n,k}^S(\psi) = \psi(\x_{n,k}) +
r_{n,k}^S$ and $\E_{n,k}(\psi) = \psi(\x_{n,k}) + r_{n,k}$,
and hence 
$$
\b_{n,k}=\[\psi(\x_{n,k}) + r_{n,k}^S\] - [\psi(\x_{n,k}) +  r_{n,k}]
= r_{n,k}^S - r_{n,k},
$$
and~\mref{bias.noise.bound.3} can be further bounded as follows:
\Array
|b_{n,k}|
&\le&
|r_{n,k}^S| + | r_{n,k}|
+ \xi_n^*\Bigl(|\psi(\x_{n,k}) + r_{n,k}^S| + |\psi(\x_{n,k}) + r_{n,k}|\Bigr)\\
&\le&
2(1+\xi_n^*)r_{n,k}^* + 2\xi_n^* |\psi(\x_{n,k})|.
\EndArray
Hence by~\mref{delta.bound.null.1}, and assumption (A4), with
probability tending to 1,
\Array
\D_n(S)
&\le&
o_p(1) +
2(1+\xi_n^*)r_n\sum_{k=1}^{K_n} W_{n,k} \Big[||\f(\x_{n,k})||_2+C_0\Big]\\
&&\qquad
+ 2\xi_n^* \sum_{k=1}^{K_n} W_{n,k} |\psi(\x_{n,k})|\\
&\le&
o_p(1) + O(r_n) + O(\xi_n^*)=o_p(1),
\EndArray
where the convergence is uniform.
\end{proof}  

\section{Limiting behavior for signal features
  (proof of Theorem~\ref{varpro.signal.theorem})}\label{appD}

The proof for $S=\{s\}$ a signal variable is similar to the noise
variable case.  The key difference is dealing with the bias term
$\b_{n,k}:=\b_{n,k}(s)$~\mref{bias.asymptotic} which is no longer asymptotically
zero.

\vskip10pt\begin{proof}
Adopting the same notation as in the proof
of~\cref{varpro.noise.theorem}, let $\E_{n,k}$ and $\E_{n,k}^s$ be
the conditional expectation for $\X$ in $R_{n,k}$ and $R_{n,k}^s$.
The same bound~\mref{psi.noise.bound} used to derive $\E_{n,k}(\psi)$
applies here.  Thus $\psi(\x) =
\psi(\x_{n,k}) + r_{n,k}(\x)$ where $|r_{n,k}(\x)|\le r_{n,k}^*$ for
$\x\in R_{n,k}$ and $\E_{n,k}(\psi) = \psi(\x_{n,k}) + r_{n,k}$ where
$r_{n,k}=\E_{n,k}(r_{n,k}(\X))\le r_{n,k}^*$.

The previous argument used for $\E_{n,k}^S(\psi)$ however no longer
applies because the released region now contains a signal variable.  To
deal with this, let $R_{n,k}=A_{n,k}\bigtimes B_{n,k}$ where
$A_{n,k}=\bigtimes_{l=1}^d I_{n,k,l}$ is the subspace of $R_{n,k}$
defined by the signal features.  By assumption (A3), $R_{n,k}$ is
shrinking to zero in the signal features, thus $I_{n,k,l}$ are
shrinking intervals for $l=1,\ldots,d$.  On the other hand,
$R_{k,n}^s$ releases the coordinates in the direction of $s$ and
therefore it is shrinking in the signal coordinates in all directions
except the $s$ direction.  This is the subspace
$A_{n,k}^s=\bigtimes_{l\in\ss\setminus s} I_{n,k,l}$.  Notice that
$A_{n,k}^s$ can be written as the union of two disjoint regions
$$
A_{n,k}^s =A_{n,k}\, \dot\bigcup\, \Acomp{s}_{n,k},\qt{where }
\Acomp{s}_{n,k}=
\bigtimes_{l=1}^{s-1} I_{n,k,l}
\bigtimes I_{n,k,s}^c
\!\!\bigtimes_{l=s+1}^{d} I_{n,k,l}
$$
and in particular this implies
\Eq
I\{\xss\in A_{n,k}^s\}
=I\{\xss\in A_{n,k}\} 
+I\{\xss\in \Acomp{s}_{n,k}\}.
\label{signal.released.regions}
\EndEq
For $\x\in R_{n,k}^s$,
$$
\psi(\x) = \big[\psi(\x_{n,k}) + r_{n,k}(\x)\big]I\{\xss\in A_{n,k}\}+
\psi(\x)I\{\xss\in \Acomp{s}_{n,k}\}.
$$
Using~\mref{signal.released.regions}, with some rearrangment, this
implies for each $\x\in R_{n,k}^s$
\Array
\psi(\x)
&=&
\psi(\x_{n,k}) I\{\xss\in A_{n,k}^s\}\\
&&\quad+
r_{n,k}(\x)I\{\xss\in A_{n,k}\}\\
&&\quad-
\[\psi(\x)-\psi(\x_{n,k})\] I\{\xss\in A_{n,k}\}\\
&& \quad+
\[\psi(\x)-\psi(\x_{n,k})\] I\{\xss\in A_{n,k}^s\}.
\EndArray
Therefore integrating with respect to $\E_{n,k}^s$,
\EqArray
\E_{n,k}^s(\psi(\X))
&=& \psi(\x_{n,k}) \nonumber\\
&&\quad + r_{n,k}^s\nonumber\\
&&\quad
- \E_{n,k}^s\Big(\[\psi(\X)-\psi(\x_{n,k})\] I\{\Xbs{\ss}\in A_{n,k}\}\Big)\nonumber\\
&&\quad
+ \E_{n,k}^s\Big(\psi(\X)-\psi(\x_{n,k})\Big)
\label{conditional.mean.signal.bound}
\EndEqArray
where $r_{n,k}^s=\E_{n,k}^s(r_{n,k}(\X) I\{\Xbs{\ss}\in A_{n,k}\})$ and
notice that
$$
|r_{n,k}^s|
\le\E_{n,k}^s(r_{n,k}^* I\{\Xbs{\ss}\in A_{n,k}\})
\le r_{n,k}^*\E_{n,k}^s(I\{\Xbs{\ss}\in A_{n,k}^s\})= r_{n,k}^*.
$$
In the proof of~\cref{varpro.noise.theorem} it was shown
$|\psi(\x)-\psi(\x_{n,k})|\le r_{n,k}^*$ for $\x\in R_{n,k}$.
Therefore the third term 
of~\mref{conditional.mean.signal.bound} is a remainder term of order
$r_{n,k}^*$.

This leaves the fourth term 
in~\mref{conditional.mean.signal.bound}.  To handle  this,
consider the local behavior of $\psi(\x)$ for $\x\in A_{n,k}^s$
around the point
$\xp_{n,k}=(x_{n,k}^{(1)},\ldots,x_{n,k}^{(s-1)},\xs{s},x_{n,k}^{(s+1)},\ldots,x_{n,k}^{(p)})'\in
A_{n,k}^s$.
By the mean-value theorem there exists 
a point $\xp_{n,k}^*=\xp_{n,k} + \l_{n,k}(\x-\xp_{n,k})$ for some
$0\le\l_{n,k}\le 1$, such that
$$
\psi(\x)-\psi(\xp_{n,k})
=(\x-\xp_{n,k})'\f(\xp_{n,k}^*).
$$
Because coordinate $s$ of $\x-\xp_{n,k}$ is zero,
\Array
|\psi(\x)-\psi(\xp_{n,k})|
&=&
\bigg|\sum_{l\in\ss\setminus s}\((\xs{l}-x_{n,k}^{(l)})\, \f^{(l)}(\xp_{n,k}^*)\)\bigg|\\
&\le&
\sum_{l\in\ss\setminus s} \(|\xs{l}-x_{n,k}^{(l)}| \cdot |\f^{(l)}(\xp_{n,k}^*)|\)\\
&\le&
\sum_{l\in\ss\setminus s} \(|\xs{l}-x_{n,k}^{(l)}| \cdot |\f^{(l)}(\xp_{n,k}^*)|\)
+ |\tilde{x}-x_{n,k}^{(s)}| \cdot |\f^{(s)}(\xp_{n,k}^*)|\\
&=&
|(\xp-\xss_{n,k})|'|\fs(\xp_{n,k}^*)|
\EndArray
where
$\xp=(x^{(1)},\ldots,x^{(s-1)},\tilde{x},x^{(s+1)},\ldots,x^{(p)})'$ and
$\tilde{x}$ can be chosen to be an arbitrary value in $I_{n,k,s}$.  Notice that $\xp\in
A_{n,k}$. Therefore the right-hand side can be bounded using the argument
of~\cref{varpro.noise.theorem} from which it follows that
$$
\Big|\psi(\x)-\psi(\xp_{n,k})\Big|\le r_{n,k}^*,
\qt{for }\x\in A_{n,k}^s.
$$
Recall that $\psi_{n,k}^s(z)=\psi(x_{n,k}^{(1)},\ldots,x_{n,k}^{(s-1)},z,x_{n,k}^{(s+1)},\ldots,x_{n,k}^d)$.
Therefore,
$\psi(\xp_{n,k})=\psi_{n,k}^s(x^{(s)})$ and
$\psi(\x_{n,k})=\psi_{n,k}^s(x_{n,k}^{(s)})$, from which it follows
\Eq
\E_{n,k}^s\Big(\psi(\X)-\psi(\x_{n,k})\Big)
=  \E_{n,k}^s\(\psi_{n,k}^s(\Xs{s})-\psi_{n,k}^s(x_{n,k}^{(s)})\)
+ O(r_{n,k}^*),
\label{conditional.mean.signal.bound.term.four}
\EndEq
and hence using~\mref{conditional.mean.signal.bound}, 
\Array
\beta_{n,k}
&=&\E_{n,k}^S(\psi) -\E_{n,k}(\psi) \\
&=& \[\psi(\x_{k,n})+ \E_{n,k}^s\(\psi_{n,k}^s(\Xs{s})-\psi_{n,k}^s(x_{n,k}^{(s)})\) 
+ O(r_{n,k}^*)\] - \Big[\psi(\x_{k,n}) + r_{n,k}\Big]\\
&=&
\E_{n,k}^s\(\psi_{n,k}^s(\Xs{s})-\psi_{n,k}^s(x_{n,k}^{(s)})\) 
+ O(r_{n,k}^*).
\EndArray
Thus the bias 
does not vanish asymptotically as in the noise
variable case.

To finish the proof we follow the rest of the proof
of~\cref{varpro.noise.theorem}. To simplify notation let
$h_{n,k}(z)=\psi_{n,k}^s(z)-\psi_{n,k}^s(x_{n,k}^{(s)})$.  
Then 
\Array
\D_n(s)
&=&
o_p(1) +
\sum_{k=1}^{K_n} W_{n,k}\Big|\b_{n,k}+\xiS{n,k}{s}\E_{n,k}^s(\psi) -
\xis{n,k}\E_{n,k}(\psi) \Big|\\
&=&
o_p(1) +
\sum_{k=1}^{K_n} W_{n,k}\Big|\b_{n,k}+\xiS{n,k}{s}\E_{n,k}^s(\psi) \Big|\\
&=&
o_p(1) +
\sum_{k=1}^{K_n} W_{n,k}\Big|(1+\xiS{n,k}{s})\E_{n,k}^s(h_{n,k}(X^{(s)})) \Big|\\
&=&
o_p(1) + (1+o_p(1))\sum_{k=1}^{K_n} W_{n,k}\Big|\E_{n,k}^s(h_{n,k}(X^{(s)})) \Big|.
\EndArray
Going from line two to line three, we have used
$\xiS{n,k}{s}\E_{n,k}^s(\psi)=\xiS{n,k}{s}\E_{n,k}^s(h_{n,k}(X^{(s)}))+o_p(1)$,
where the $o_p(1)$ term is uniform and is due
to~\mref{conditional.mean.signal.bound} combined
with~\mref{conditional.mean.signal.bound.term.four}.
Finally, the last line holds because
\Array
&&\hskip-45pt
(1+\xi_n^*)\sum_{k=1}^{K_n} W_{n,k}\Big|\E_{n,k}^s(h_{n,k}(X^{(s)})) \Big|\\
&&\ge
\sum_{k=1}^{K_n} W_{n,k}\Big|(1+\xiS{n,k}{s})\E_{n,k}^s(h_{n,k}(X^{(s)})) \Big|\\
&&\ge
\sum_{k=1}^{K_n} W_{n,k}\Big|\E_{n,k}^s(h_{n,k}(X^{(s)})) \Big|-
 \xi_{n}^* \sum_{k=1}^{K_n} W_{n,k}\Big|\E_{n,k}^s(h_{n,k}(X^{(s)}))\Big|\\
&&= 
(1-\xi_n^*)\sum_{k=1}^{K_n} W_{n,k}\Big|\E_{n,k}^s(h_{n,k}(X^{(s)})) \Big|.
 \EndArray
The right inequality is because $|a+b|\ge |a|-|b|$ for any
real-valued $a, b$.
\end{proof}

\section{Regression synthetic experiments used for benchmarking}\label{appE}

Regression simulation models used to test VarPro are
listed below:

\Enumerate
\setlength\itemsep{1pt}

\item {\it cobra2}:
  $\psi(\x)=\xs{1}\xs{2}+(\xs{3})^2-\xs{4}\xs{7}+\xs{8}\xs{10}-(\xs{6})^2$,
  $\Xs{j}\sim U(-1, 1)$,
  $\e\sim N(0, 0.1^2)$.



\item {\it cobra8}:
  $\psi(\x,\e)=I\{\xs{1}+(\xs{4})^3+\xs{9}+\sin(\xs{2}\xs{8})+\e>0.38\}$,
  $\Xs{j}\sim U(-.25, 1)$, $\e\sim N(0, 0.1^2)$.

\item {\it friedman1}:
  $\psi(\x)= 10 \sin(\pi \xs{1}  \xs{2}) + 20  (\xs{3} - 0.5)^2 + 10  \xs{4} + 5  \xs{5}$,
  $\Xs{j}\sim U(0, 1)$, $\e\sim N(0, 1)$.


\item {\it friedman3}:
  $\psi(\x)= \displaystyle{\arctan\[\frac{\xs{2} \xs{3} - 1/(\xs{2}\xs{4})}{\xs{1}}\]}$,
  $\Xs{1}\sim U(0,100)$, $\Xs{2}\sim U(40\pi,560\pi)$,
  $\Xs{3},\ldots,\Xs{p}\sim U(0,1)$, $\e\sim N(0, 1)$.
  
\item {\it inx1}: $\psi(\x)=\xs{1}(\xs{2})^2\sqrt{|\xs{3}|}+\lfloor
  \xs{4}-\xs{5}\xs{6}\rfloor$, $\Xs{j}\sim U(-1, 1)$, $\e\sim N(0, 0.1^2)$.

\item {\it inx2}:
  $\psi(\x)=\xs{3}(\xs{1}+1)^{|\xs{2}|}-\sqrt{\dfrac{(\xs{5})^2}{|\xs{4}|+|\xs{5}|+|\xs{6}|}}$,
  $\Xs{j}\sim U(-1, 1)$,
  $\e\sim N(0, 0.1^2)$.

\item {\it inx3}:
  $\psi(\x)=\cos(\xs{1}-\xs{2})+\arcsin(\xs{1}\xs{3})-\arctan(\xs{2}-(\xs{3})^2)$,
  $\Xs{j}\sim U(-1, 1)$,
  $\e\sim N(0, 0.1^2)$.

\item {\it lm}:
  $\psi(\x)=\sum_{j=1}^{15} \xs{j}$, $\Xs{j}\sim N(0,1)$, $\e\sim N(0,15^2)$.

\item {\it lmi1}:
  $\psi(\x)=.05 f_1(\x) + \exp(.02 f_1(\x) f_2(\x))$, where
  $f_1(\x)=\sum_{j=1}^{10} \xs{j}$, $f_2(\x)=\sum_{j=11}^{20} \xs{j}$,
  $\Xs{j}\sim U(0,1)$, $\e\sim N(0,.1^2)$.

\item {\it lmi2}:
  $\psi(\x)=3(\sum_{j=1}^{15} \xs{j})^2$, 
  $\Xs{j}\sim N(0,1)$, $\e\sim N(0,15^2)$.


  
\item {\it sup}: $\psi(\x)=10 \xs{1}\xs{2}+.25\dfrac{1}{\xs{3}\xs{4}+10
  \xs{5}\xs{6}}$, $\Xs{j}\sim U(0.05, 1)$, $\e\sim N(0, 0.5^2)$.

\item {\it sup2}:
  $\psi(\x)=\pi^{\xs{1}\xs{2}}\sqrt{2 \xs{3}}-\arcsin(\xs{4})
      +\log(\xs{3}+\xs{5})-\dfrac{\xs{9}}{\xs{10}}\sqrt{\dfrac{\xs{7}}{\xs{8}}}-\xs{2}\xs{7}$,
      $\Xs{j}\sim U(0.5, 1)$, $\e\sim N(0, 0.5^2)$.

\item {\it supX}: \hskip2pt Same as {\it sup} but with smaller $N$ and
  larger $p$.

\item {\it supX2}: \hskip2pt Same as {\it sup2} but with smaller $N$ and larger $p$.

\EndEnumerate

Simulations {\it cobra} are from~\cite{biau2016cobra}
and {\it friedman} are from~\cite{friedman1991multivariate}.
All experiments used $N=2000$ and
$p=40$ except for {\it supX} and {\it sup2X} where $N=500$ and $p=200$.
In a first set of runs, features
were independently sampled as described above.  In a second
run, all features retained the same marginal distribution as
before, but were transformed using a copula so as to make all
features correlated with correlation $\rho=0.9$.  This was done for
all simulations except {\it lm} and {\it lmi2} where
the 15 signal features
$\Xs{1},\ldots,\Xs{15}$ were correlated within blocks
of size 5 (1--5, 6--10 and 11--15).

\begin{figure}[phtb]
  \renewcommand\thefigure{E1}
  \centering
\resizebox{6in}{!}{\includegraphics[page=1]{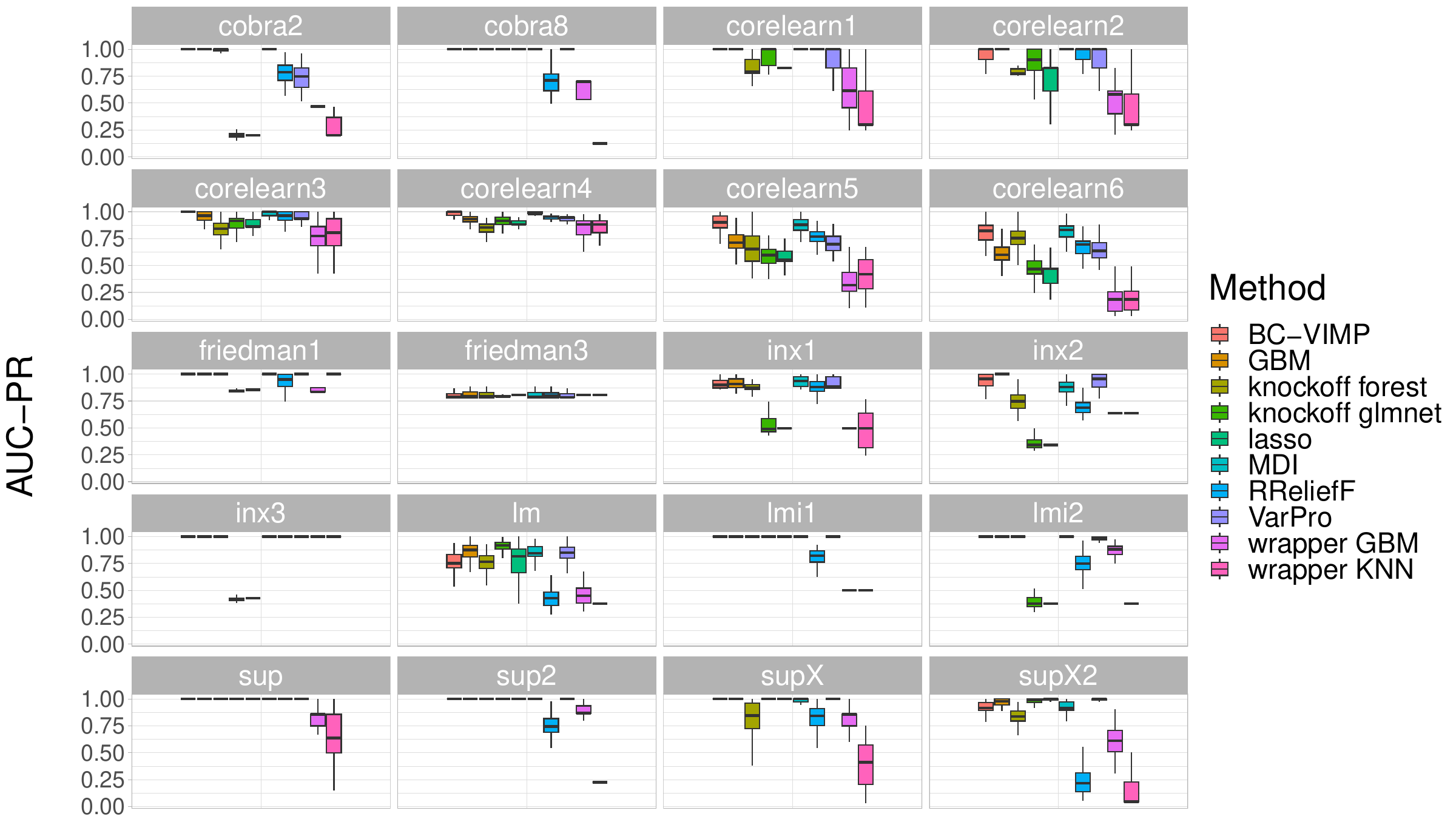}}\vskip20pt
\resizebox{6in}{!}{\includegraphics[page=2]{benchmark.pdf}}
\vskip10pt
  \caption{\small \label{regTab} Area under the precision recall curve
    (AUC-PR) and gmean (geometric mean of TPR and TNR) feature
    selection performance in 
    regression simulations where variables are uncorrelated.}
\label{benchmark.uncorrelated}
\end{figure}

\begin{figure}[phtb]
  \renewcommand\thefigure{E2}
\centering
\resizebox{6in}{!}{\includegraphics[page=3]{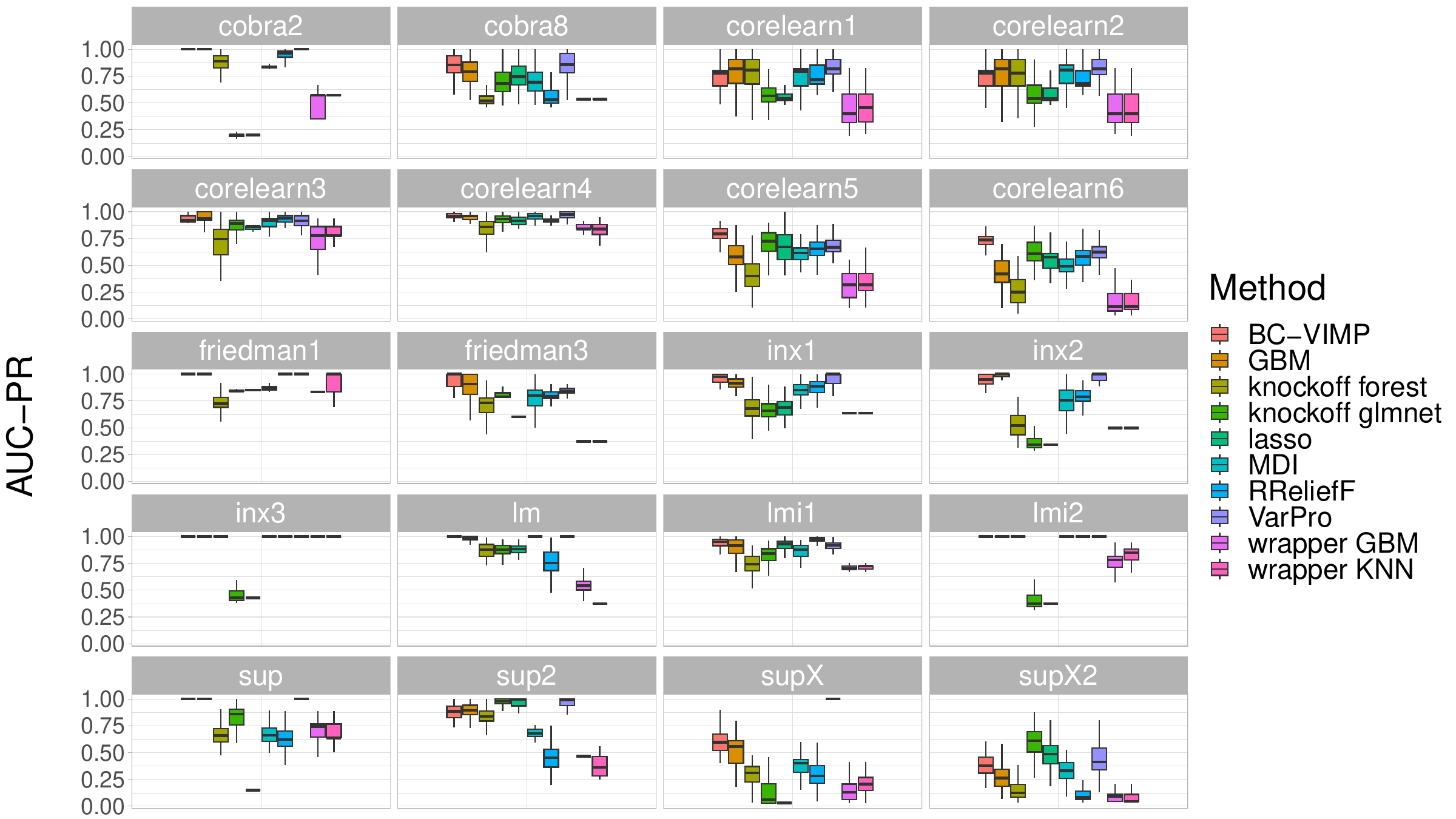}}\vskip20pt
\resizebox{6in}{!}{\includegraphics[page=4]{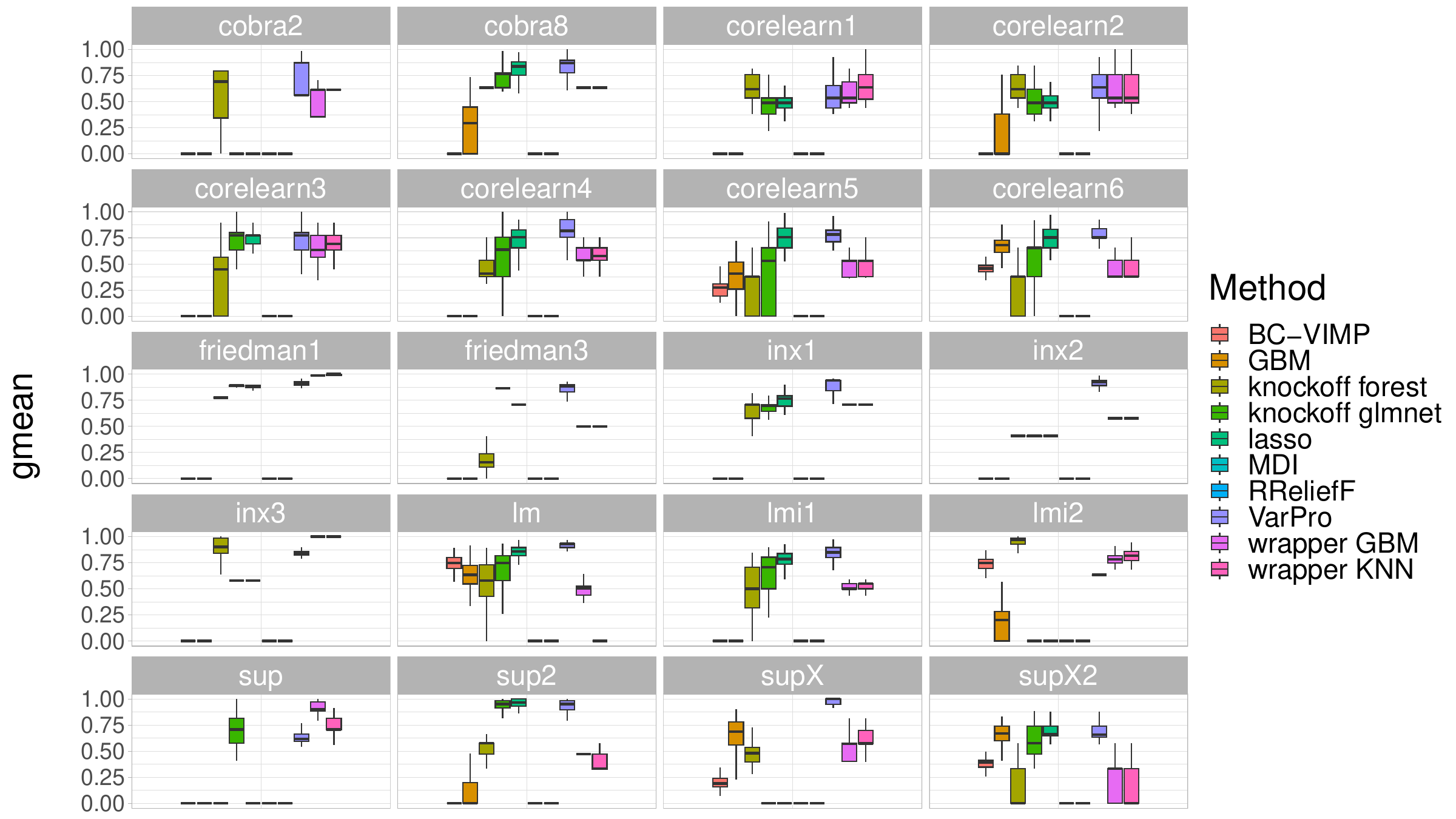}}
\vskip10pt
\caption{\small Similar to~\cref{benchmark.uncorrelated} but using
  correlated variables.}
\label{benchmark.correlated}
\end{figure}

\subsection{Additional simulations}
Synthetic data was also obtained using
the function {\ttfamily regDataGen} from the R-package \core.
In these experiments, the
response is randomly selected from two different functions. The choice
depends on a ``switch variable'' which determines whether
$\psi(\x)=\xs{4}-2\xs{5}+3\xs{6}$ or
$\psi(\x)=\cos(4\pi\xs{4})(2\xs{5}-3\xs{6})$.  Therefore, half the
time the model is linear and half the time it is nonlinear.
Also there are 4 discrete variables $a_1,a_2,a_3,a_4$, the first two
$a_1,a_2$ carry information about the
switch variable.  Variables $\Xs{1},\Xs{2}$ also
carry information about the hidden variable.
The simulation was modified to allow for additional noise
variables drawn from a $U(0,1)$ distribution.  Different $N$ and $p$
and varying configurations for information about the hidden variable
were used.  The following simulations were considered:

\Enumerate
\setlength\itemsep{2pt}
\item[15.] {\it corelearn1}:
$a_1,a_2$ and $\Xs{1},\Xs{2}$ contain no information; $N=300$.
\item[16.] {\it corelearn2}:
$a_1,a_2$ and $\Xs{1},\Xs{2}$ contain no information; $N=200$.
\item[17.] {\it corelearn3}:
$a_1,a_2$ contain full information and $\Xs{1},\Xs{2}$ contain no information; $N=100$.
\item[18.] {\it corelearn4}:
$a_1,a_2$  and $\Xs{1},\Xs{2}$ contain full information; $N=100$.
\item[19.] {\it corelearn5}:
$a_1,a_2$  and $\Xs{1},\Xs{2}$ contain full information; $N=100$; 50 noise
  variables added.
\item[20.] {\it corelearn6}:
$a_1,a_2$  and $\Xs{1},\Xs{2}$ contain full information; $N=100$; 200 noise
  variables added.
\EndEnumerate

In the first set of runs, features were sampled as described in
{\ttfamily regDataGen}.  In the second run, all features retained the
same marginal distribution as before, but were transformed to have
correlation $\rho=0.9$.  Note that signal variables for the
first two simulations are $\Xs{4},\Xs{5},\Xs{6}$ while the third
simulation adds $a_1,a_2$ and the last three simulations
adds $a_1,a_2$ and $\Xs{1},\Xs{2}$ as 
signals.

\section{Asymptotics for the modified
procedure (proof of {Theorem~\ref{varpro.external.noise.theorem}})}\label{appF}

The following assumption will be used in the proof:
\Enumerate
\setlength\itemsep{5pt}
\item[(A5)]
There exists a sequence $\rp_n\rightarrow 0$ and subspace
$\xx_n\subseteq\xx$ containing all regions 
$\bigcup_{k=1}^{K_n}R_{k,n}$ and released regions $\bigcup_{k=1}^{K_n}R_{n,k}^{S}$
such that $|\psi_n(\x)-\psi(\x)|\le  \rp_n$ for $\x\in\xx_n$.
\EndEnumerate
The assumption requires $\psi_n$ to converge uniformly to
$\psi$ but some flexibility is allowed in that convergence only has
to hold over a suitably defined subspace.  For example, if
$\psi_n(\x)=h(\sum_{l=1}^p\alpha_{n,l}\xs{l})$ and
$\psi(\x)=h(\sum_{l=1}^p\alpha_{0,l}\xs{l})$ for $h$ a real-valued
function with derivative $h'$, then by the mean value theorem
\Array
|\psi_n(\x)-\psi(\x)|
&\le&
\Bigg|h'\(\sum_{l=1}^p\alpha_{n,l}^*\xs{l}\)\Bigg|
\sum_{l=1}^p\Big|\xs{l}\(\alpha_{n,l}-\alpha_{0,l}\)\Big|\\
&\le&
\Bigg|h'\(\sum_{l=1}^p\alpha_{n,l}^*\xs{l}\)\Bigg|
\cdot ||\x||_2 \cdot \sqrt{\sum_{l=1}^p|\alpha_{n,l}-\alpha_{0,l}|^2}
\EndArray
where $\alpha_{n,l}^*$ is some value between $\alpha_{n,l}$ and
$\alpha_{0,l}$.  The simplest way to satisfy (A5) is to require
boundedness where $\xx_n\subseteq\xx_0$ for $\xx_0$ a closed bounded
subspace of $\xx$.  Then (A5) holds under the relatively mild
assumptions that $h'$ is continuous and
$\sum_{l=1}^p|\alpha_{n,l}-\alpha_{0,l}|\rightarrow 0$ where
convergence can be at any rate.  The boundedness condition is easily
met as the size of a region is entirely controlled by the data
analyst.  

\vskip10pt\begin{proof}
For the proof we use a centering argument for $\Ds_n(S)$ similar to
that used for $\D_n(S)$.  Let
$Z_i^\star(\z)=\psi(\X_i)I\{\X_i\in R(\z)\}-b(\z)$ where
$b(\z)=\E[\psi(\X)I\{\X\in R(\z)\}]$.
Using $\psi_n=\psi+(\psi_n-\psi)$, it follows that
\EqArray
\ts_n(\z_{n,k}^S)-
\ts_n(\z_{n,k})|
&=&
\[\frac{1}{m_{n,k}^S}\sum_{i=1}^n Z_i^\star(\z_{n,k}^S)
- \frac{1}{m_{n,k}}\sum_{i=1}^n Z_i^\star(\z_{n,k})\]
\nonumber\\
&&+
\[\frac{n b(\z_{n,k}^S)}{m_{n,k}^S}-\frac{n b(\z_{n,k})}{m_{n,k}}\]
\nonumber\\
&&+
\[\frac{1}{m_{n,k}^S}\sum_{i=1}^n \Zs_{n,i}(\z_{n,k}^S)
- \frac{1}{m_{n,k}}\sum_{i=1}^n \Zs_{n,i}(\z_{n,k})\],
\label{expansion.mod}
\EndEqArray
where $\Zs_{n,i}(\z)=[\psi_n(\X_i)-\psi(\X_i)]I\{\X_i\in R(\z)\}$.
Observe that the second term in~\mref{expansion.mod}
is the bias term asymptotically equal to $\b_{n,k}(S)$~\mref{bias.asymptotic} 
worked out in the previous theorems.
The terms in the first square bracket
in~\mref{expansion.mod} are sums of i.i.d.\ centered variables
and therefore are similar to the sums in~\mref{expansion} and can be
dealt with by~\cref{key.lemma} to show they converge to zero
uniformly in probability.  Therefore we only need consider the terms
inside the third
backet of~\mref{expansion.mod}.

Therefore, consider the bound
\EqArray
&&\sum_{k=1}^{K_n} W_{n,k}\Bigg|
\frac{1}{m_{n,k}^S}\sum_{i=1}^n \Zs_{n,i}(\z_{n,k}^S)
- \frac{1}{m_{n,k}}\sum_{i=1}^n \Zs_{n,i}(\z_{n,k})\Bigg|\nonumber\\
&&\qquad\le
\sum_{k=1}^{K_n}\frac{W_{n,k}}{m_{n,k}^S}\sum_{i=1}^n |\Zs_{n,i}(\z_{n,k}^S)|
+\sum_{k=1}^{K_n}\frac{W_{n,k}}{m_{n,k}}\sum_{i=1}^n
|\Zs_{n,i}(\z_{n,k})|.
\label{external.bound}
\EndEqArray
Begin with the first sum on the right of~\mref{external.bound}.
By (A5),
$$
|\Zs_{n,i}(\z_{n,k}^S)|
\le |\psi_n(\X_i)-\psi(\X_i)|  I\{\X_i\in R_{n,k}^S\}
\le \rp_n  I\{\X_i\in R_{n,k}^S\}.
$$
Therefore
$$
\sum_{k=1}^{K_n}\frac{W_{n,k}}{m_{n,k}^S}\sum_{i=1}^n |\Zs_{n,i}(\z_{n,k}^S)|
\le r_n\sum_{k=1}^{K_n}\frac{W_{n,k}}{m_{n,k}^S}\sum_{i=1}^n I\{\X_i\in R_{n,k}^S\}
=r_n\rightarrow 0.
$$
The second sum on the right of~\mref{external.bound} involving $\Zs_{n,i}(\z_{n,k})$ is dealt with
similarly.
\end{proof}

\end{appendix}



\vskip20pt
\section*{Code Availability}
Our code is publicly available as an
R-package \varPro\, and is available at the repository
\url{https://github.com/kogalur/varPro}.

\begin{funding}
Research for the authors was supported by the National Institute Of
General Medical Sciences of the National Institutes of Health, Award
Number R35 GM139659 and the National Heart, Lung, and Blood Institute
of the National Institutes of Health, Award Number R01 HL164405.
\end{funding}

\bibliographystyle{imsart-number} 
\bibliography{varPro}       

\end{document}